\documentclass[10pt,twocolumn,letterpaper]{article}

\usepackage{cvpr}
\usepackage{times}
\usepackage{epsfig}
\usepackage{graphicx}
\usepackage{amsmath}
\usepackage{amssymb}
\usepackage{mathtools}
\usepackage[normalem]{ulem}
\usepackage{algorithmic}
\usepackage{algorithm}
\usepackage{epstopdf}
\usepackage{subcaption}
\usepackage{color}
\usepackage{epstopdf}
\usepackage{authblk}
\DeclareMathOperator*{\argmin}{{\arg\!\min}}

\newlength\myindent
\newcommand\bindent{%
        \begingroup
        \setlength{\itemindent}{\myindent}
        \addtolength{\algorithmicindent}{\myindent}
}
\newcommand\eindent{\endgroup}

\newcommand{\p}{\mathbf{p}}
\newcommand{\tr}{\mathbf{t}}
\newcommand{\vv}{\mathbf{v}}
\newcommand{\diag}{\mathrm{diag}}
\newcommand{\rank}{\mathrm{rank}}
\newcommand{\trace}{\mathrm{trace}}
\newcommand{\spann}{\mathrm{span}}

\newcommand{\comment}[1]{{ }}

\usepackage[pagebackref=true,breaklinks=true,letterpaper=true,colorlinks,bookmarks=false]{hyperref}

 \cvprfinalcopy 

\def\cvprPaperID{1992} 

\ifcvprfinal\fi
\begin{document}

\title{A New Rank Constraint on Multi-view Fundamental Matrices,\\
and its Application to Camera Location Recovery}

\author[1]{Soumyadip Sengupta} 
\author[2]{Tal Amir}
\author[2]{Meirav  Galun}
\author[4] {Tom Goldstein}
\author[1]{David W. Jacobs} 
\author[3]{Amit Singer}
\author[2]{Ronen  Basri}
\affil[1]{Center for Automation Research, University of Maryland, College Park, USA.} 
\affil[2]{Department of Computer Science and Applied Mathematics, Weizmann Institute of Science, Israel.}
\affil[3]{Department of Mathematics and PACM, Princeton University, USA.}
\affil[4]{Department of Computer Science, University of Maryland, College Park, USA.}

\maketitle

\begin{abstract}
Accurate estimation of camera matrices is an important step in structure from motion algorithms. In this paper we introduce a novel rank constraint on collections of fundamental matrices in multi-view settings. We show that in general, with the selection of proper scale factors, a matrix formed by stacking fundamental matrices between pairs of images has rank 6. Moreover, this matrix forms the symmetric part of a rank 3 matrix whose factors relate directly to the corresponding camera matrices. We use this new characterization to produce better estimations of fundamental matrices by optimizing an L1-cost function using Iterative Re-weighted Least Squares and Alternate Direction Method of Multiplier. We further show that this procedure can improve the recovery of camera locations, particularly in multi-view settings in which fewer images are available.
\end{abstract}

\section{Introduction}
Accurate reconstruction of 3D scenes from multiview stereo images is one of the primary goals of computer vision. Current techniques use point correspondences to estimate either the essential or fundamental matrices between pairs of images, and then use the estimated matrices to recover the camera matrices and structure. Notable success was achieved when sequential methods were introduced~\cite{agarwal2009building,snavely2006photo}. These methods first recover camera matrices and structure from two images. Then, adding one image at a time, they apply bundle adjustment to estimate the camera matrix (and structure) of the new image. Recent work attempts to further improve recovery by considering simultaneously subsets of multiple images and recovering camera matrices that are consistent over the entire subsets.  Indeed a number of papers have focused on the consistent recovery of \textit{either} camera orientation or location~\cite{arie2012global,ozyesil2015stable,ozyesil2015robust,tron2014distributed,wilson2014robust,martinec2007robust}.


\begin{figure}[t]
  \centering
 \includegraphics[width=.24\textwidth]{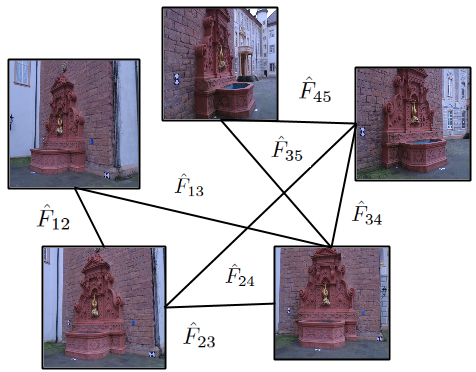}
 \\[2.5mm]
\footnotesize{\tiny
$\underbrace{\left[ \hspace{-1mm}\begin{smallmatrix}
\mathbf{0} & \hat{F}_{12} & \hat{F}_{13} & - & - \\
  \hat{F}_{21} & \mathbf{0} & \hat{F}_{23} & \hat{F}_{24} &  - \\
  \hat{F}_{31} & \hat{F}_{32} & \mathbf{0} &  \hat{F}_{34} & \hat{F}_{35}\\
  - & \hat{F}_{42} & \hat{F}_{43} & \mathbf{0} & \hat{F}_{45} \\
  - & - & \hat{F}_{53} & \hat{F}_{54} & \mathbf{0}\\
\end{smallmatrix} \hspace{-1mm} \right]}
\approx
\underbrace{ \color{red}{ \left[ \begin{smallmatrix}
  \mathbf{0} & \lambda_{12} & \lambda_{13} & \lambda_{14} & \lambda_{15} \\[0.3mm]
  \lambda_{21} & \mathbf{0} & \lambda_{23} & \lambda_{24} & \lambda_{25} \\[0.3mm]
  \lambda_{31} & \lambda_{32} & \mathbf{0} & \lambda_{34} & \lambda_{35} \\[0.3mm]
  \lambda_{41} & \lambda_{42} & \lambda_{43} & \mathbf{0} & \lambda_{45} \\[0.3mm]
  \lambda_{51} & \lambda_{52} & \lambda_{53} & \lambda_{54} & \mathbf{0}
    \end{smallmatrix} \right]}}
    \odot
    \underbrace{\left[ \begin{smallmatrix}
  \mathbf{0} & F_{12} & F_{13} & \color{red}{F_{14}} & \color{red}{F_{15}}\\[0.3mm]
  F_{21} & \mathbf{0} & F_{23} & F_{24} &  \color{red}{F_{25}} \\[0.3mm]
  F_{31} & F_{32} & \mathbf{0} & F_{34} & F_{35} \\[0.3mm]
  \color{red}{F_{41}} & F_{42} & F_{43} & \mathbf{0} & F_{45}\\[0.3mm]
  \color{red}{F_{51}} & \color{red}{F_{52}} & F_{53} & F_{54} & \mathbf{0}
    \end{smallmatrix} \right]}
$\\ \hspace*{-1.8mm}$\hat F$ \hspace{26mm} $\Lambda$ \hspace{26mm} $F$ \\[0.5mm]
\uline{with $F=A+A^T$ and $\rank(A)=3$}.}
  \caption{\small Illustration of our rank constraint. Collections of fundamental matrices $\{\hat F_{ij}\}$ estimated for pairs of images (top) are arranged in a matrix $\hat F$ (bottom). This matrix should be equal (up to noise) to a matrix $F$ or properly scaled fundamental matrix, which in turn forms the symmetric part of a rank 3 matrix $A$.}
  \label{fig:overall}
\end{figure}

This paper introduces new constraints to enable the consistent recovery of fundamental and essential matrices. This is potentially advantageous since those matrices capture simultaneously the location and orientation of the cameras, along (in the case of fundamental matrices) with their internal calibration parameters. For configurations of cameras that are not all collinear, our main result establishes that, when scaled properly, the matrix formed by appending all pairwise fundamental matrices in a multiview setting is of rank 6. More tightly, this matrix forms the symmetric part of a rank 3 matrix whose factors relate directly to the entries of the corresponding camera matrices. We further show that multiview settings of collinear cameras yield a rank 4 matrix.

We use this characterization to develop an optimization formulation for estimating consistent sets of fundamental matrices. Our formulation can accept sets of estimated fundamental matrices in which some are noisy, some are outliers, and some cannot be estimated at all from image pairs (i.e., missing data). In solving this optimization we seek a set of scaled fundamental matrices that satisfy our constraints and fit the estimated fundamental matrices. Our formulation uses an L1 cost function, which is optimized with Iterative Re-weighted Least Squares (IRLS) \cite{holland1977robust}, to remove outliers, and uses Alternate Direction Method of Multipliers (ADMM) \cite{boyd2011distributed} to incorporate rank constraints.

Our work is related to a variety of approaches to structure from motion (SfM) that utilize rank constraints. Tomasi and Kanade showed that under an orthographic projection, and after centering, projected points form a rank 3 matrix. Sturm and Triggs~\cite{sturm1996factorization,triggs1996factorization} extended this to perspective projection by showing that projected points, when scaled properly, form a rank 4 matrix. Unlike their work, which uses rank constraint on tracks of points in images, our work only considers fundamental matrices and so in multiview settings it gives rise to systems with many fewer variables, relying on potentially less noisy estimates. Our approach, which seeks to recover a consistent set of fundamental matrices, is analogous to rotation or translation averaging and to loop closure \cite{hartley2013rotation,chatterjee2013efficient,cui2015global}. In fact, obtaining consistent fundamental matrices can be regarded as simultaneous averaging of rotation, translation and camera calibration and as a way to close all loops. Our experiments indicate that such joint averaging performs better than a separate averaging of rotation and translation.

A number of algorithms have recently been proposed for solving unconstrained,  low rank systems with outliers and missing data (e.g., \cite{candes2009exact,hu2013fast,okatani2007wiberg}) with remarkable success. Extending such techniques to incorporate SfM constraints is an important next step.

When thousands of images are available, existing methods that use pairwise epipolar constraints or tri-focal tensors can exploit highly over-determined systems to handle noise and outliers quite accurately. However, when fewer images are available the importance of rank constraints grows, and their introduction can potentially yield more accurate estimation of camera parameters. Indeed, we provide experiments that show that using our characterization, essential matrices can be estimated more accurately than with current state-of-the-art methods, and these in turn can be translated to better estimates of camera locations.


\comment{
\subsection{Relation to Prior Work}

Rank constraints have been widely used in computer vision \cite{tomasi1992shape,torresani2001tracking,cui2012background,cheng2013rank}. Altough rank constraints are non-convex, researchers have developed techniques to handle them based in explicit factorization \cite{okatani2007wiberg} or a convex relaxation \cite{candes2009exact,hu2013fast}.  \cite{tomasi1992shape} showed that the matrix containing a collection of corresponding points across images (termed the measurement matrix), is rank 4 under orthographic projection . For perspective projection \cite{triggs1996factorization} extended this idea to a rank 4 decomposition of a scaled measurement matrix, where scales denote projective depth. Matrix completion for Structure from Motion is studied in the literature in  \cite{kennedy2016online,fragkiadaki2014grouping,cabral2013unifying,jacobs1997linear}. \cite{dai2010element} developed a semidefinite programming based algorithm to perform low rank decomposition in the presence of missing data and outliers. These methods have been less successful in recovering camera parameters for unconstrained real world images, as they heavily rely on point correspondences, which are very noisy and contain large amounts of outliers. On the other hand current methods in SfM (both global and iterative) robustly estimate pair-wise camera geometry, e.g.,~ the fundamental matrix, from point correspondences in images, followed by recovery of camera locations and orientations. This two step process helps to reduce the noise and outliers effectively for unconstrained real world images. This motivates us to develop robust low rank methods to improve estimates of pair-wise fundamental matrices that are initially estimated from point correspondences of pairs of images.\\

We now describe the current state-of-the-art pipeline used in global SfM and then explain how we can use our method  to improve the pipeline.
The first step of the process is to estimate corresponding feature points between pairs of images and recover the fundamental matrices relating them. The second step involves recovering camera parameters (rotation and translation) from these pair-wise estimations. Lastly alternate minimization of the camera parameters and the 3D scene is performed using Bundle Adjustment \cite{triggs1999bundle}, either using incremental (\cite{agarwal2009building,wu2011visualsfm,snavely2006photo,zhang2001incremental}) or global methods \cite{goldstein2016shapefit,wilson2014robust,ozyesil2015robust,ozyesil2015stable}.  With global methods the problem of estimating camera orientation (rotation) and location (translation) are done separately. Current methods can efficiently solve the orientation estimation problem; recent work focuses on solving the more challenging problem of estimating camera location.  \cite{tron2016survey} provides a nice review of all these techniques.\\

Image pairs provide the relative direction of two cameras.
 In \cite{arie2012global} the problem of combining these to find camera locations is solved using a least square formulation. Tron \textit{et al.} \cite{tron2014distributed} developed a constrained least squares based formulation to solve for camera locations. In \cite{wilson2014robust}, Wilson \textit{et al.} have used a pre-processing technique, based on projection in many random directions, to get rid of outliers in the original pair-wise direction measurements. Ozyesil \textit{et al.}\cite{ozyesil2015stable} introduced a semidefinite relaxation based SDR solver using additional non-convex constraints along with a least squares formulation. In \cite{ozyesil2015robust}, Ozyesil \textit{et al.} formulated the location recovery problem as an L1 optimization and solved it using IRLS.
In \cite{goldstein2016shapefit}, Goldstein \textit{et al.} have shown a theoretical guarantee on the recovery of camera locations and solved the same problem using an ADMM formulation.\\

In this work, we use the rank constraint to improve camera location estimation. To estimate camera rotation we follow \cite{ozyesil2015robust}, which uses an iterative estimation method based on \cite{arie2012global}. Using the estimated rotations for each camera and the fundamental matrices estimated by our method we obtain pair-wise camera directions. We feed these into \cite{ozyesil2015robust}'s algorithm for camera locatoin recovery, improving its performance.
 In \cite{ozyesil2015robust,ozyesil2015stable} and \cite{goldstein2016shapefit} the initial camera directions are obtained from pairs of images only, whereas our method performs matrix completion and produces pair-wise directions using global information. A summary of our method is shown in Figure \ref{fig:pipeline} and is discussed in detail in Section \ref{sec:exp}.

}

\section{Low-Rank Characterization of Fundamental Matrices in Multiview Settings}
\label{sec:low-rank}

\subsection{Background}
We first introduce notations and give a short summary of the relevant concepts in multi-view geometry. An extensive discussion of this topic can be found in~\cite{hartley2003multiple}. Let $I_{1},...,I_{n}$ denote a collection of $n$ images of a scene and let $\tr_{i} \in \mathbb{R}^{3}$ and  $R_{i} \in SO(3)$ denote the location and  orientation of the  $i$'th camera in a global coordinate system. Let the $3 \times 3$ $K_i$ denote the intrinsic camera calibration matrix for $I_i$. $K_i$ is nonsingular and is typically specified in the form
\begin{equation}
K_i = \begin{bmatrix}
f_x & \alpha & u_0 \\ 0 & f_y & v_0 \\ 0 & 0 & 1
\end{bmatrix},
\end{equation}
where, $f_x$ and $f_y$ respectively are the focal lengths in the $x$ and $y$ direction, $(u_0,v_0)$ form the principal point and $\alpha$ represents the skew coefficient.
Let $P = (X,Y,Z)^T$ be a scene point in the global coordinate system. Its projection onto $I_i$ (expressed in homogeneous coordinates) is given by $\p_i = P_i/Z_i$, where $P_i = (X_i,Y_i,Z_i)^T = K_i R^{T}_i(P-\tr_i)$. We therefore associate with $I_i$ the $3 \times 4$ camera matrix $C_i = K_i R^T_i \begin{bmatrix}I, -\tr_i\end{bmatrix}$, where $I$ is a $3 \times 3$ identity matrix and note that scaling $C_i$ does not affect projection.

Next, we consider the relations between pairs of images, $I_i$ and $I_j$. We can express the camera rotation and translation relating two images by $R_{ij}=R^{T}_i R_j$ and $\tr_{ij}=R^{T}_i(\tr_i -\tr_j)$. Clearly, $R_{ji}=R^{T}_{ij}$ and $\tr_{ji}=-R^{T}_{ij}\tr_{ij}$. Two images are further related by epipolar line constraints, which are expressed by $\p^{T}_i F_{ij} \p_j = 0$, where $F_{ij}$ denotes the fundamental matrix relating $I_i$ to $I_j$. $F_{ij}$ can be estimated up to scale from point correspondences. $F_{ij}$ is related to the rotation and translation between $I_i$ and $I_j$ and to their respective calibration matrices by $F_{ij} = K^{-T}_i [\tr_{ij}]_{\times} R_{ij} K^{-1}_j$, where $[\tr_{ij}]_{\times}$ denotes the skew-symmetric matrix corresponding to cross-product with $\tr_{ij}$. In cases in which the cameras are calibrated we set $K_i=K_j=I$ and replace the fundamental matrix with the essential matrix $E_{ij}=[\tr_{ij}]_{\times} R_{ij}$. Therefore, $F_{ij}=K^{-T}_iE_{ij}K^{-1}_j$.

To derive our rank constraint we will need to express the essential and fundamental matrices relative to a global coordinate system. \cite{zhang1997general} derived an expression in terms of the camera matrices $C_i$ and $C_j$. Here we will use the more recent derivation of~\cite{arie2012global} that, as we shall see below, is amenable to factorization:
\begin{align}
E_{ij} =& R^{T}_i (T_i-T_j) R_j, \label{eq:baseE} \\
F_{ij} =& K^{-T}_i R^{T}_i (T_i-T_j) R_j K^{-1}_j,
\label{eq:baseF}
\end{align}
where $T_i = [\tr_i]_\times$.

\subsection{Low-rank Construction}
\label{section:low-rank}

We next introduce our main result, which includes a low rank characterization of the collection of fundamental matrices in multiview settings. For our result we will construct a matrix of size $3n \times 3n$, denoted $F$, in which each of the $3 \times 3$ blocks includes a fundamental matrix $F_{ij}$ (see Figure~\ref{fig:overall}), where we assume that each of the pairwise fundamental matrices in $F$ is scaled properly. We further define $F_{ii}=0$ for all $1 \le i \le n$, and note that this is consistent with~\eqref{eq:baseF}. Likewise we define the $3n \times 3n$ matrix $E$ from the essential matrices $E_{ij}$. We refer to $F$ (resp.~$E$) as the \textit{multiview matrix of fundamentals (essentials)}.

\medskip

\noindent\textbf{Claim 1}: $F$ (and likewise $E$) is symmetric and $\rank(F) \le 6$. Moreover,
\begin{enumerate}
\item If $F$ is produced by $n$ cameras whose centers are not all collinear then $\rank(F)=6$ and there exists a $3n \times 3n$ matrix $A$ with $\rank(A)=3$ such that $F=A+A^T$.
\item If $F$ is produced by $n$ cameras whose centers are all collinear then $\rank(F) \le 4$ and there exists a matrix $A$ with $\rank(A) \le 2$ such that $F=A+A^T$.
\end{enumerate}

\medskip

\noindent\textbf{Proof}: To prove the claim we begin by defining the matrix $A$ as follows. Let  $U_i = K^{-T}_i R^{T}_i T_i$, $V_i = K^{-T}_i  R^{T}_i$, and $A_{ij}=U_iV_j^T$. $U_i$, $V_i$, and $A_{ij}$ are $3 \times 3$ matrices. Observing~\eqref{eq:baseF} and recalling that $T_i$ is skew-symmetric we see that $F_{ij}=A_{ij}+A_{ji}^T$.

Next we construct the $3n \times 3$ matrices $U$ and $V$ as
\[ U = \begin{bmatrix} U_1\\ \vdots\\ U_n \end{bmatrix} ~ \mathrm{and} ~~
V = \begin{bmatrix} V_1\\ \vdots\\ V_n \end{bmatrix}\]
and set $A=UV^T$. Clearly, by construction, $\rank(A) \le 3$. Moreover, $F=A+A^T$, and so $F$ is symmetric and $\rank(F) \le 6$.

\smallskip

\noindent\textbf{\uline{Case 1}}: We show next that unless the cameras are all collinear $\rank(A)=3$. Clearly $\rank(V)=3$. Therefore we need to show that also $\rank(U)=3$. We prove this by contradiction. Assume $\rank(U) < 3$. Then $\exists\,\tr \in \mathbb{R}^3$, $\tr \ne \boldsymbol0$, s.t.\ $U\tr=\boldsymbol0$. This implies that $\tr_i \times \tr = \boldsymbol0$ for all $1 \le i \le n$. Thus, all the $\tr_i$'s are parallel to $\tr$, violating our assumption that not all camera locations are collinear. Consequently $\rank(U)=3$ and therefore also $\rank(A)=3$.

Next we show that when the cameras are not all collinear $\rank(F)=6$. We recall that $F_{ij}=K^{-T}_iE_{ij}K^{-1}_j$ where $K_i$ and $K_j$ are non-singular. We can therefore write $F=K^TEK$ where the $3n \times 3n$ matrix $K$ is block diagonal with blocks formed by $\{K_i^{-1}\}_{i=1}^n$. This implies that $\rank(F)=\rank(E)$, and so we are left to show that $\rank(E)=6$.

We assume WLOG that the camera locations are centered at the origin, i.e., $\sum_{i=1}^{n} \tr_i = 0$ (since $E$ is invariant to global translation of the cameras). We further argue that each column of $U$ is orthogonal to each column of $V$. This is evident from the following identities
\begin{align}
V^{T} U &= \sum_{i=1}^{n} V^{T}_i U_i =
\sum_{i=1}^{n} T_i = \left[\sum_{i=1}^{n} \tr_i\right]_{\times} = 0_{3 \times 3}.
\end{align}
Let $\tilde A$ denote the matrix $A$ where we substitute $K_i=I, \forall i$ (so that $E=\tilde A+\tilde A^T$.) Denote by $\tilde A = \hat{U} \Sigma \hat{V}^T$ the SVD of $\tilde A$ ($\hat{U}$ and $\hat{V}$ are $3n \times 3$ and $\Sigma$ is $3 \times 3$). Since $\tilde A=U V^T$ we have that $\spann(U) = \spann(\hat{U})$ and $\spann(V) = \spann(\hat{V})$. Now we can decompose $E$ as :
\begin{align}
E &= \tilde A + \tilde A^T = \hat{U} \Sigma \hat{V}^T + \hat{V} \Sigma \hat{U}^T \nonumber \\
&=  \begin{bmatrix} \hat{U} \hspace{2mm} \hat{V} \end{bmatrix}  \begin{bmatrix}
\Sigma \hspace{4mm}\\ \hspace{4mm} \Sigma
\end{bmatrix}\begin{bmatrix} \hat{V}^T \\ \hat{U}^T \end{bmatrix}
\label{eq:Edeco}
\end{align}
Since the columns of $U$ are orthogonal to those of $V$, the matrix $\begin{bmatrix} \hat{U} \hspace{2mm} \hat{V} \end{bmatrix}$ is column orthogonal. Thus, \eqref{eq:Edeco} is the SVD of $E$. And since $\tilde A$ is rank 3, $\Sigma$ is full rank. Consequently, $\rank(F)=\rank(E)=6$.

\smallskip

\noindent\textbf{\uline{Case 2}}: Suppose all camera centers are collinear. WLOG assume that the origin of the global coordinate system is also collinear with the $n$ cameras (since $F$ is unaffected by global translation), and so we can write $\tr_i=\alpha_i\tr$ for $1 \le i \le n$ where $\alpha_i \in \mathbb{R}$ and $\tr \in \mathbb{R}^3$. Let $T=[\tr]_\times$, then clearly $U_i=\alpha_i K^{-T}_i R^{T}_i T$. Define $\tilde U_i=\alpha_iK^{-T}_i R^{T}_i$ (so $U_i = \tilde U_i T$) and let the $3n \times 3$ matrix $\tilde U$ be formed by stacking $U_1, U2,...$ on top of each other then
\[ A = U V^T = \tilde U T V^T. \]
Since $T$ is skew-symmetric its rank is at most 2 and so is $\rank(A)$. It follows that $\rank(F) \leq 4$. $\blacksquare$


\subsection{Tightness of our constraints}

Claim 1 provides two constraints on the $3n \times 3n$ matrix $F$.
\begin{itemize}
  \item $F=A+A^T$ and $\rank(A)=3$.
  \item The diagonal block of $F$ vanishes, i.e., $F_{ii}=0$.
\end{itemize}

We now investigate how tight these constraints are in producing fundamental matrices that are consistent with a set of camera parameters. We show that the number of degrees of freedom allowed by these constraints is equal to the number of degrees of freedom in the camera matrices. However, we find that there exist matrices that are allowed by these constraints, but do not produce valid fundamental matrices.


Counting arguments show that our constraints allow $12n-15$ degrees of freedom (DOFs) in defining $F$. Specifically, since $A$ is rank 3 it can be written as $A=UV^T$ where $U$ and $V$ are $3n \times 3$, so together they have $18n$ entries. The constraint $F=A+A^T$, however, gives rise to a 15 DOF ambiguity that should be subtracted from the number of entries of $U$ and $V$, as we explain in the next paragraph. The constraint that $F_{ii}=0$ requires $U_i V_i^T$ to be skew symmetric, yielding $6n$ more constraints on the entries of $U$ and $V$, yielding together $12n-15$ DOFs.

To calculate the DOFs in the ambiguity of $F=A+A^T$ note that we can write $F$ as $F=[U,V]J[U,V]^T$, where $J$ is a $6 \times 6$ permutation matrix defined as $J=\begin{bmatrix}\mathbf{0}&I\\I&\mathbf{0}\end{bmatrix}$ (so $J[U,V]^T=[V,U]^T$). With this notation the ambiguity in factorizing $F$ is obtained by introducing a $6 \times 6$ matrix Q such that $QJQ^T=J$ so that $[U,V]QJQ^T[U,V]^T=[U,V]J[U,V]^T=F$. $Q$ has 36 entries, but the constraints $QJQ^T=J$ reduces its degrees of freedom to 15. Denote $Q=\begin{bmatrix}Q_{11}&Q_{12}\\Q_{21}&Q_{22}\end{bmatrix}$ these constraints restrict the products $Q_{11}Q_{12}$ and $Q_{21}Q_{22}$ to be skew symmetric and the sum $Q_{11}Q_{22}+Q_{12}Q_{21}=I$, providing altogether 21 constraints on the 36 entries of $Q$, leaving 15 DOFs.

Coincidentally, the number of DOFs in factoring $F$ is equal to the DOFs in defining $n$ cameras. In general, the number of DOFs in defining $n$ perspective cameras is $11n-15$. However, each camera matrix can be scaled arbitrarily and each choice of scale will (inversely) scale the respective row and column of  $F$. In other words, $n$ camera matrices, $C_1,...,C_n$, scaled arbitrarily by non zeros $1/s_1,...,1/s_n$, produce a collection of equivalent multiview fundamental matrices defined by
\[\{SFS|S=\diag\{s_1,s_1,s_1,s_2,...,s_n\},s_{i} \ne 0\}.\]
The freedom in choosing the entries of $S$ accounts for the $n$ missing DOFs.

We note however that although the DOFs in factoring $F$ with our constraints are equal to the DOFs in defining $n$ camera matrices there exist matrices that satisfy our constraints but cannot be realized with $n$ cameras. Specifically, these constraints do not guarantee that all the pairwise fundamental matrices $F_{ij}$ are rank deficient. The constraint $F_{ii}=0$ restricts $U_iV_i^T$ to be skew-symmetric, implying that either $U_i$ or $V_i$ is rank deficient. If all $U_i$'s (or equivalently all $V_i$'s) are chosen to be rank deficient then so are all the $F_{ij}$. If however some of the $U_i$'s and some of the $V_i$'s are chosen to be full rank then they may produce $F_{ij}$ blocks that are rank 3 and so they are not legal fundamental matrices. Note that the skew-symmetry of $U_i V_i$ guarantees that no more than 1/4 of the $F_{ij}$'s can be of full rank. Indeed, our experiments (in Section \ref{sec:exp}) often produce $F_{ij}$'s that are near rank 2; in a typical run the average ratio of the third to second largest singular value ≈ $7 \times 10^{-8}$ , presumably because the problem is so over-constrained.

In conclusion, while our constraints provide a necessary but not sufficient conditions for consistency, counting considerations indicate that our constraints are nearly tight. Below we develop and optimization scheme that utilizes these constraints to infer the missing scale factors for collections of estimated pairwise fundamental matrices, to recover missing fundamentals and to correct noisy ones.

\section{Low-rank Constrained Optimization to Recover Fundamental Matrices}
\label{sec:opti}
In this section we formulate an optimization problem that uses the constraints derived in section \ref{sec:low-rank} to achieve a better recovery of pairwise fundamental matrices. Assume we are given a set of fundamental matrices $\hat{F}_{ij}$, where $(i,j) \in \Omega$ and $\Omega$ denotes the subset of image pairs for which fundamental matrices  have been estimated. (We will further assume $(i,j) \in \Omega \implies (j,i) \in \Omega$.) We use these matrices to construct our measurement matrix $\hat{F}$ whose $(i,j)$'s $3 \times 3$ block contains $\hat{F}_{ij}$ if $(i,j) \in \Omega$ and is zero otherwise. Note that in the absence of errors each non-zero block is related by an unknown scale factor $\lambda_{ij}$ to the corresponding block in the sought multiview matrix of fundamentals $F$, where $\lambda_{ij}$ depends on the distance between the $i$'th and $j$'th cameras. Recovering these scale factors is essential in order to apply our constraints. Our task therefore can be expressed as:
\begin{align}  \label{eq:opt}
\min_{F,\{\lambda_{ij}\}} \sum \limits_{(i,j) \in \Omega} \|\hat{F}_{ij} - \lambda_{ij} F_{ij} \|_{Fro},
\end{align}
where $F$ is constrained to fulfill the constraints in Claim 1. Here we have chosen to minimize over the sum of Frobenius norms of each $3 \times 3$ block. Such mixed L1-L2 norm minimization is expected to be robust to outlier estimates of $F_{ij}$'s.

We note that the formulation~\eqref{eq:opt} is bilinear in $F$ and the scale factors. We could avoid this bilinearity by minimizing instead $\|\lambda_{ij} \hat{F}_{ij} - F_{ij} \|_{Fro}$. Such minimization, however, is subject to a zero trivial solution and so it requires an additional constraint such as $\sum_{ij} \lambda_{ij}^2=1$. Our experience with such formulation is that it is quite sensitive to errors.

Expressing~\eqref{eq:opt} with the constraints results in the following problem:
\begin{align}
\min_{A,\{\lambda_{ij}\}} & \quad \frac{1}{2}\sum \limits_{(i,j) \in \Omega} \|\hat{F}_{ij} - \lambda_{ij} (A_{ij} + A_{ji}^{T})\|_{Fro} \nonumber \\
\mathrm{s.t.} & \quad \rank(A)=3, ~ A_{ii}+A^{T}_{ii}=0, ~ \lambda_{ij}=\lambda_{ji}
\label{opti:main}
\end{align}
where $A_{ij}$ denotes each $3 \times 3$ sub-block of $A$. Our solution for $F$ then is $F=A+A^T$.

\eqref{opti:main} introduces a number of challenges, including the mixed L1-Frobenius norms, the bilinearity,  and the rank constraint. This problem is non-convex due to the latter two challenges. Below we describe how we approach these challenges with IRLS and ADMM. Our algorithm is summarized in Algorithm \ref{algo:our}.

\subsection{Handling Outliers with IRLS}
We begin by addressing the mixed L1-Frobenius norm in the cost function. We approach this with Iterative Re-weighted Least Squares (IRLS) \cite{holland1977robust}. IRLS converts the problem to weighted least squares where the weights are updated from one iteration to the next. At each iteration $t$ of the IRLS we replace the cost function in~\eqref{opti:main} with
\begin{align}
\min_{A,\{\lambda_{ij}\}} \frac{1}{2} \sum_{(i,j) \in \Omega} w^t_{ij} \|\hat{F}_{ij} - (A_{ij} + A_{ji}^{T}) \lambda_{ij}\|^2_{Fro},
\label{opti:wls}
\end{align}
where
\begin{align}
w^{t}_{ij} = \begin{dcases}
1/\max(\delta,\|\hat{F}_{ij} - \lambda^{t-1}_{ij}(A^{t-1}_{ij} + (A^{t-1}_{ji})^T)\|_{Fro}), \\
   \hspace{16mm} \textrm{if} ~ (i,j) \in \Omega\\
0  \hspace{15mm} \textrm{otherwise.} \end{dcases}  \nonumber
\end{align}
$\delta$ is a regularization parameters (we use $\delta=10^{-3}$).

 To clarify presentation we simplify our notations as follows. Let $W$ and $\Lambda$ be $3n \times 3n$ matrices. Denoting their $3 \times 3$ sub-blocks by $W_{ij}$ and $\Lambda_{ij}$, we set $W_{ij} = w_{ij} \mathbf{1}$ and $\Lambda_{ij} = \lambda_{ij} \mathbf{1}$, where $\mathbf{1}$ is a $3 \times 3$ matrix with all 1's. We further use the subscript $WF$ to denote the weighted Frobenius norm, i.e., $\|\vv\|^2_{WF} = \trace(\vv^TW\vv)$ and use $\odot$ to denote element-wise product of matrices. Therefore, in each IRLS iteration we seek to solve
\begin{align}  \label{opti:full}
\underset{A,\Lambda}{\min} & \quad \frac{1}{2}\|\hat{F} - \Lambda \odot (A+A^T) \|^2_{WF}  \\
\mathrm{s.t.} & \quad \rank(A)=3, ~A_{ii} + A_{ii}^T = 0, ~\Lambda_{ij} = \lambda_{ij} \mathbf{1}, ~\lambda_{ij}=\lambda_{ji}. \nonumber
\end{align}

\subsection{Optimization using ADMM}
Next, we wish to solve the non-convex optimization problem in \eqref{opti:full}, including the bilinearity and the rank constraint. To this end we will use a scaled version of Alternate Direction Method of Multiplier (ADMM)~\cite{boyd2011distributed,goldstein2014fast}. We maintain a second copy of $A$, which we denote as $B$, and form the augmented Lagrangian of (\ref{opti:full}) as:
\begin{align}
&\underset{\Gamma}{\max} ~ \underset{A,B,\Lambda}{\min} \quad \frac{1}{2}\|\hat{F} - \Lambda \odot (A+A^T) \|^2_{WF} + \frac{\tau}{2}\|B-A+\Gamma\|^2_{Fro} \nonumber \\
&\mathrm{s.t.} ~ \rank(B)=3, ~A_{ii} + A_{ii}^T = 0, ~\Lambda_{ij} = \lambda_{ij} \mathbf{1}, ~\lambda_{ij}=\lambda_{ji}.
\label{opti:admm}
\end{align}
The last term in this objective, $\frac{\tau}{2} \|B-A+ \Gamma \|^2_F$ denotes the Lagrangian penalty; $\tau$ is a constant, and $\Gamma$ is a matrix of Lagrange multipliers of the same size as $A$ that is updated in the ADMM steps. We next describe the ADMM steps, which are applied iteratively.

\bigskip

\noindent
\textbf{\uline{Step 1}: Solving for $(A,\Lambda)$}.\\
In each iteration, $k$, we solve the following sub-problems:
\begin{align}
&\underset{A,\Lambda}{\min} \quad \frac{1}{2}\|\hat{F} - \Lambda \odot (A+A^T) \|^2_{WF} + \frac{\tau}{2}\|A-(B+\Gamma)\|^2_{Fro} \nonumber \\
&\mathrm{s.t.} ~A_{ii} + A_{ii}^T = 0, ~\Lambda_{ij} = \lambda_{ij} \mathbf{1}, ~\lambda_{ij}=\lambda_{ji}.
\label{opti:A}
\end{align}
Since \eqref{opti:A} is non-convex we will solve it by alternative minimization of $A$ and $\Lambda$
\begin{enumerate}
\item Optimize w.r.t. $A$:\\
Because of the form of~\eqref{opti:A} it is useful to separate $A$ into its symmetric and anti-symmetric parts, $A_s$ and $A_n$, so that $A=\frac{1}{2}(A_s + A_n)$ with $A_s = A+ A^T$ and $A_n = A-A^T$. Let $G= B+\Gamma$; $G_s$ and $G_n$ respectively denote its symmetric and anti-symmetric part. We can write (\ref{opti:A}) in terms of $A_s$ and $A_n$ and separately solve for them as follows:
\begin{align}
A^{(k+1)}_s = & ~\underset{A_s}{\argmin} ~ \frac{1}{2}\|\hat{F} - \Lambda^{(k)} \odot A_s \|^2_{WF} \nonumber \\
 + & ~\frac{\tau}{8}\|A_s - G_s^{(k)}\|^2_F ~~ \mathrm{s.t.} ~~  (A_s)_{ii} = 0, \label{opti:As} \\
A^{(k+1)}_n = & ~\underset{A_n}{\argmin} ~ \frac{\tau}{8}\|A_n - G_n^{(k)}\|^2_F
= ~ G^{(k)}_n.
\label{eq:An}
\end{align}
To solve (\ref{opti:As}) we take the derivative w.r.t.~$A_s$ and equate to 0. Thus we update $A_s$ according to
\begin{align}
A^{(k+1)}_s = & ~W \odot \Lambda^{(k)} \odot \hat{F} + \frac{\tau}{4} G^{(k)}_s
\\  \oslash & ~(W \odot \Lambda^{(k)} \odot \Lambda^{(k)} + \frac{\tau}{4})
\nonumber
\\ (A^{(k+1)}_s)_{ii} = & ~0
\label{eq:As}
\end{align}
where $\oslash$ denotes element-wise division.

\item Optimize w.r.t.~$\Lambda$:
We minimize the following sub-problem
\begin{align}
\Lambda^{(k+1)} = & ~\underset{\Lambda}{\argmin} ~ \|\hat{F} - \Lambda \odot A^{(k+1)}_s\|^2_{WF} \nonumber \\
\mathrm{s.t.} ~ & ~\Lambda_{ij} = \lambda_{ij} \mathbf{1}, ~\lambda_{ij}=\lambda_{ji}.
\label{opti:L}
\end{align}
We can solve (\ref{opti:L}) separately for each block as follows,
\begin{align}
\lambda^{(k+1)}_{ij} &= \underset{\lambda_{ij}}\argmin ~ \|\hat{F}_{ij} - \lambda_{ij} (A^{(k+1)}_s)_{ij} \|^2_{WF}, ~~ i < j \nonumber \\
&=\frac{ \trace(\hat{F}^T_{ij}(A^{(k+1)}_s)_{ij} ) }{\|(A^{(k+1)}_s)_{ij}\|^2_{Fro}}
\label{eq:L}
\end{align}
Note that $\lambda^{(k+1)}_{ii} = 0$, $\lambda^{(k+1)}_{ji} = \lambda^{(k+1)}_{ij}$ and $\Lambda^{(k+1)}_{ij} = \lambda^{(k+1)}_{ij} \mathbf{1}$.

\end{enumerate}

\medskip

\noindent
\textbf{\uline{Step 2}: Solving for $B$}.\\
This part of the ADMM deals with the rank constraint. It requires a solution to

\begin{align}
B^{(k+1)} &= \underset{B}\argmin ~ \frac{\tau}{2} ||B - A^{(k+1)} + \Gamma^{(k)}||_{Fro}^2 \nonumber \\
 &\mathrm{s.t.} ~ \rank(B)=3.
\end{align}
This is solved by
\begin{align}
B^{(k+1)}&= SVP(A^{(k+1)} - \Gamma^{(k)},3),
\label{eq:B}
\end{align}where $SVP(X,r)$ denotes the Singular Value Projection (SVP) of X into space the of rank-$r$ matrices. To perform $SVP(X,r)$ we compute the SVD of $X$ and keep its top $r$ singular values and the corresponding singular vectors.

\bigskip

\noindent
\textbf{\uline{Step 3}: Update of $\Gamma$.}
The matrix $\Gamma$ contains Lagrange multipliers that are used in the saddle-point formulation~\eqref{opti:admm} to enforce the equality constraint $A=B$.    The following update is a gradient ascent step that acts to maximize the augmented Lagrangian \eqref{opti:admm} for $\Gamma.$  For details, see \cite{boyd2011distributed,goldstein2014fast}.
\begin{eqnarray}
\Gamma^{(k+1)} = \Gamma^{(k)} + (B^{(k+1)} - A^{(k+1)}).
\label{eq:G}
\end{eqnarray}

\begin{algorithm}[!h]
   \caption{IRLS-ADMM solver}
   \label{algo:our}
   \begin{algorithmic}
      \STATE \textbf{Input:} Estimated fundamentals in $\hat{F}$ and $\Omega$.
      \STATE \textbf{Output:} Recovered $F$.
      \STATE \textbf{IRLS:} Solve \eqref{opti:main}
      \STATE Initialize $\Lambda$ and $A$.
.      \STATE Create weights for IRLS, $w^0_{ij} = 1$ if $(i,j) \in \Omega$ and $w^0_{ij} = 0$ otherwise. Set $t=1$.

      \WHILE{not converged}
      \STATE Solve \eqref{opti:wls} using ADMM formulation \eqref{opti:admm}.
      \bindent
      \STATE Set $k=0$, $\tau=\sum w_{ij}$, $\Gamma^0 = 0$. $B=A.$
      \WHILE{not converged}
      \STATE Alternative minimization of \eqref{opti:A}.
      \bindent
      \STATE Update $A$ using \eqref{eq:An} and \eqref{eq:As}.
      \STATE Update $\Lambda$ using \eqref{eq:L}.
      \eindent
      \STATE Update $B$ using \eqref{eq:B}
.      \STATE Update $\Gamma$ using \eqref{eq:G}
.      \STATE $k=k+1$.
      \ENDWHILE
      \eindent
      \STATE Update Weights $w^{t}_{ij}$ using \eqref{opti:wls}.
      \STATE $t=t+1$.
      \ENDWHILE
      \STATE $F=A+A^T$.
   \end{algorithmic}
\end{algorithm}

\begin{figure}[!h]
	\centering
	\includegraphics[width=.35\textwidth]{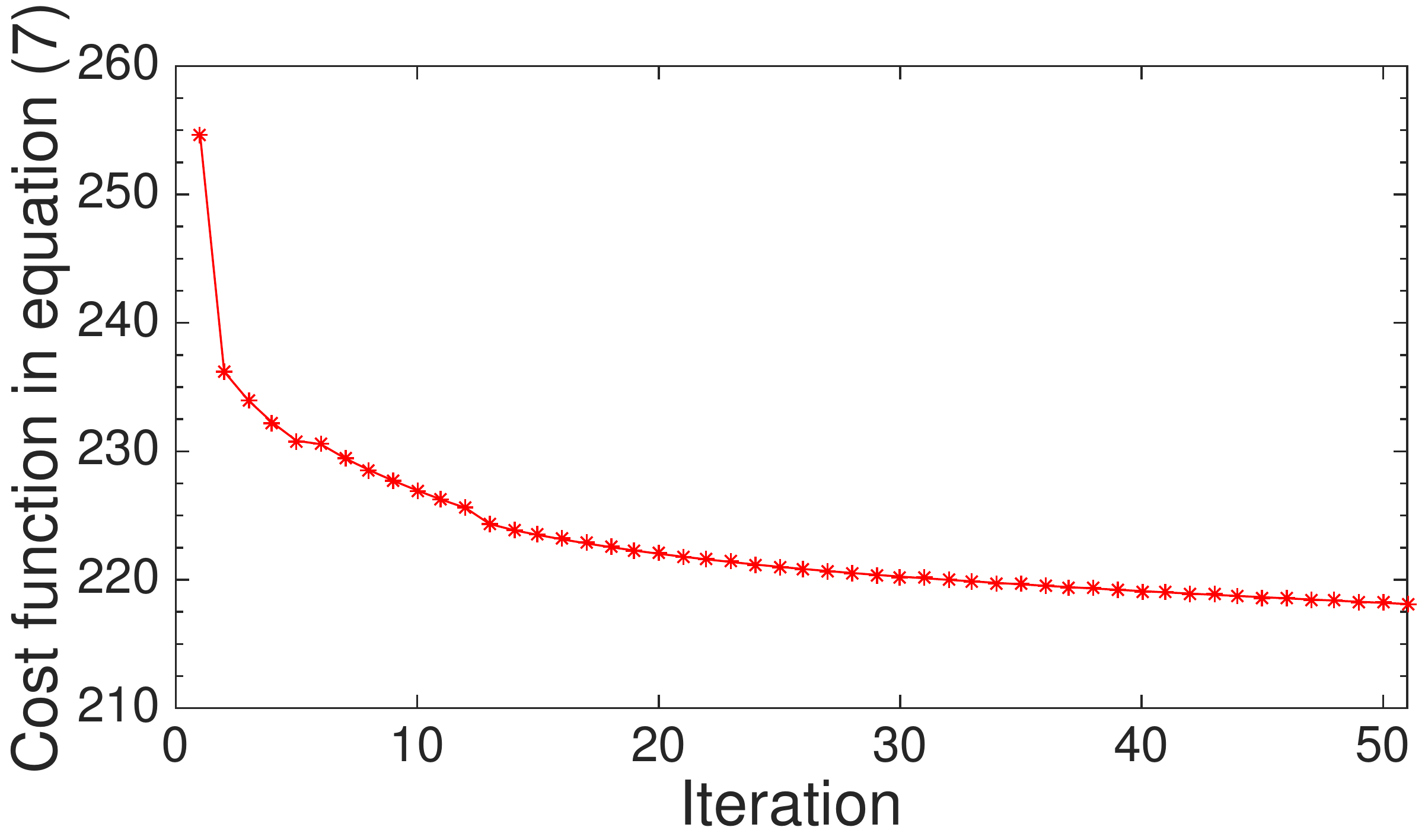}
	\caption{\small Convergence of our optimization algorithm}
	\label{fig:conv}
\end{figure}

Empirically we observe monotonic convergence of the cost function defined in Equation \ref{opti:main} with each iteration of IRLS on a sample problem as shown in Figure \ref{fig:conv}. For every iteration of the IRLS we run ADMM till convergence to optimize Equation \ref{opti:admm}.



\begin{figure}[!h]
        \centering
        \includegraphics[width=.24\textwidth]{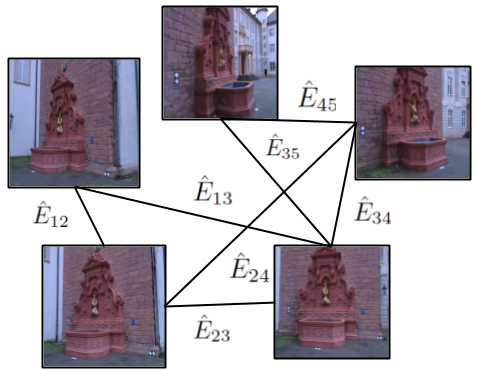}
        \includegraphics[width=.3\textwidth]{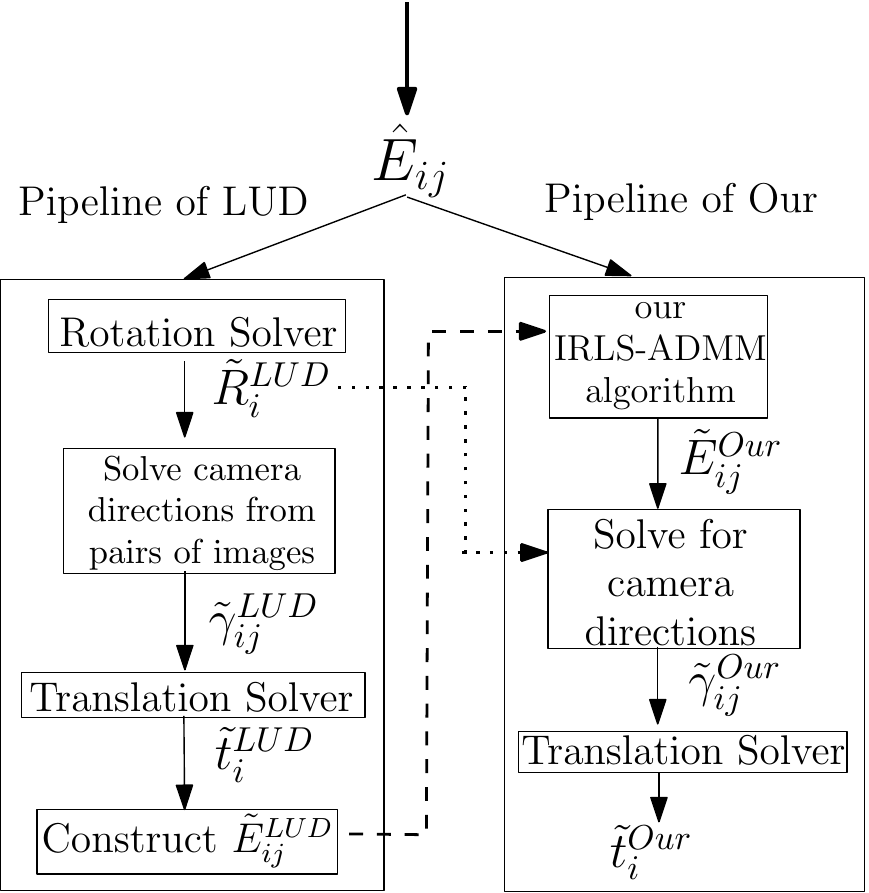}
                \caption{\small SfM pipelines for LUD (left) and our method (right).}
        \label{fig:pipeline}
\end{figure}

\bigskip

\section{Experiments}
\label{sec:exp}

To demonstrate the utility of our method we tested it in the problem of estimating essential matrices and camera locations from multiple images. Current iterative and global approaches to Structure from Motion (SfM) are often tested on large datasets when many pairwise essential matrices can be estimated, achieving outstanding performance. We argue that imposing rank constraints  can be useful particularly when the number of images is relatively small. To demonstrate this we run our method on subsets of images of different sizes showing improved performance relative to the existing methods particularly with smaller subsets.

\comment{We focus our experiments on essential matrices.} In many common SfM pipelines the intrinsic calibration parameters are recovered separately. Therefore, while our method can be applied when the calibration parameters are unknown, here we assume that the cameras are calibrated and so we apply our optimization algorithm to essential matrices. Note that our derivations in Sections \ref{sec:low-rank} and \ref{sec:opti} hold also for essential matrices by setting $K_i =I$.

We next describe the tested methods:
\begin{itemize}
\item \textbf{LUD \cite{ozyesil2015robust}}: Figure \ref{fig:pipeline} shows the pipeline used by LUD to estimate camera locations and orientations from pairs of images. Starting from pairwise essential matrices estimated with SIFT \cite{lowe2004distinctive} and RANSAC \cite{bolles1981ransac}, this method first solves for camera orientations, denoted by $\tilde{R}^{\mathrm{LUD}}_i$ in Figure \ref{fig:pipeline}, by iteratively applying \cite{arie2012global} while rejecting outliers. Using camera orientations it then returns to the image keypoints to estimate pairwise camera directions, denoted by $\tilde{\gamma}^{\mathrm{LUD}}_{ij}$. Using these pairwise directions it applies IRLS to solve for camera locations ($\tilde{t}^{\mathrm{LUD}}_{i}$), which we compare to our method. In addition, we use the estimated camera locations and orientations to reconstruct the pairwise essential matrices $\tilde{E}^{\mathrm{LUD}}_{ij}$.
\item \textbf{ShapeKick \cite{goldstein2016shapefit}}: For this method we use the same pipeline as used with LUD, except that we replace the translation recovery part of LUD with ShapeKick. ShapeKick formulates the location recovery problem as a convex optimization and solves it with ADMM. They achieved comparable performance to LUD on the dataset of \cite{wilson2014robust}.
\item \textbf{1DSfM \cite{wilson2014robust}}: This method uses a pre-processing technique, based on projection in many random directions, to remove outliers in the original pairwise direction measurements. In our experiments we use their software, which uses the pipeline described in \cite{wilson2014robust} and only provides camera locations.
\item \textbf{Our method}: Figure \ref{fig:pipeline} shows the pipeline used by our method. From the pair-wise essential matrices we minimize \eqref{opti:main} using the IRLS-ADMM summarized in Algorithm \ref{algo:our}. Since our method is not convex it requires a good initialization. We initialize it with essential matrices produced by the LUD method of Ozyesil \textit{et al.} \cite{ozyesil2015robust},  denoted $\tilde{E}^{\mathrm{LUD}}_{ij}$. Specifically $\tilde{E}^{LUD}_{ij}$ is used to initialize $\Lambda$ and $A$ in Algorithm \ref{algo:our}. Our algorithm improves these essential matrix estimates, producing a collection of new pairwise estimates in $E$, denoted $\tilde{E}^{\mathrm{Our}}_{ij}$. To further produce camera locations we first use $\tilde{E}^{\mathrm{Our}}_{ij}$  and the rotations obtained by the LUD pipeline, $\tilde{R}^{\mathrm{LUD}}_i$, to solve for the pairwise camera directions $\tilde{\gamma}^{\mathrm{Our}}_{ij}$. Then we apply translation solver of LUD to the $\tilde{\gamma}^{\mathrm{Our}}_{ij}$ with $(i,j) \in \Omega$ to produce camera locations $\tilde{t}^{\mathrm{Our}}_{i}$. As is shown below, our improved estimates of essential matrices lead in turn to improved estimates of camera locations compared to the LUD pipeline.
\end{itemize}

We tested these methods on real image collections from \cite{wilson2014robust}, which comes with `ground truth' estimates of camera locations and essential matrices produced with a sequential method similar to \cite{snavely2006photo}. (These ground truth estimates are used also in \cite{wilson2014robust,ozyesil2015robust,goldstein2016shapefit}.) For our experiments we used 14 different scenes from the dataset. For each scene we randomly selected 5 different sub-samples of $N$ images from the dataset. We used $N=50$, 100, and 150 images, resulting in 70 different trials for each $N$. In each trial we compared the quality of the essential matrix recovered by our method to that recovered by LUD and ShapeKick. Likewise, we compared the quality of our recovered camera locations to those obtained by the three competing methods.

\begin{figure}[!h]
	\centering
	\begin{subfigure}{0.48\textwidth}
		\includegraphics[width=\textwidth]{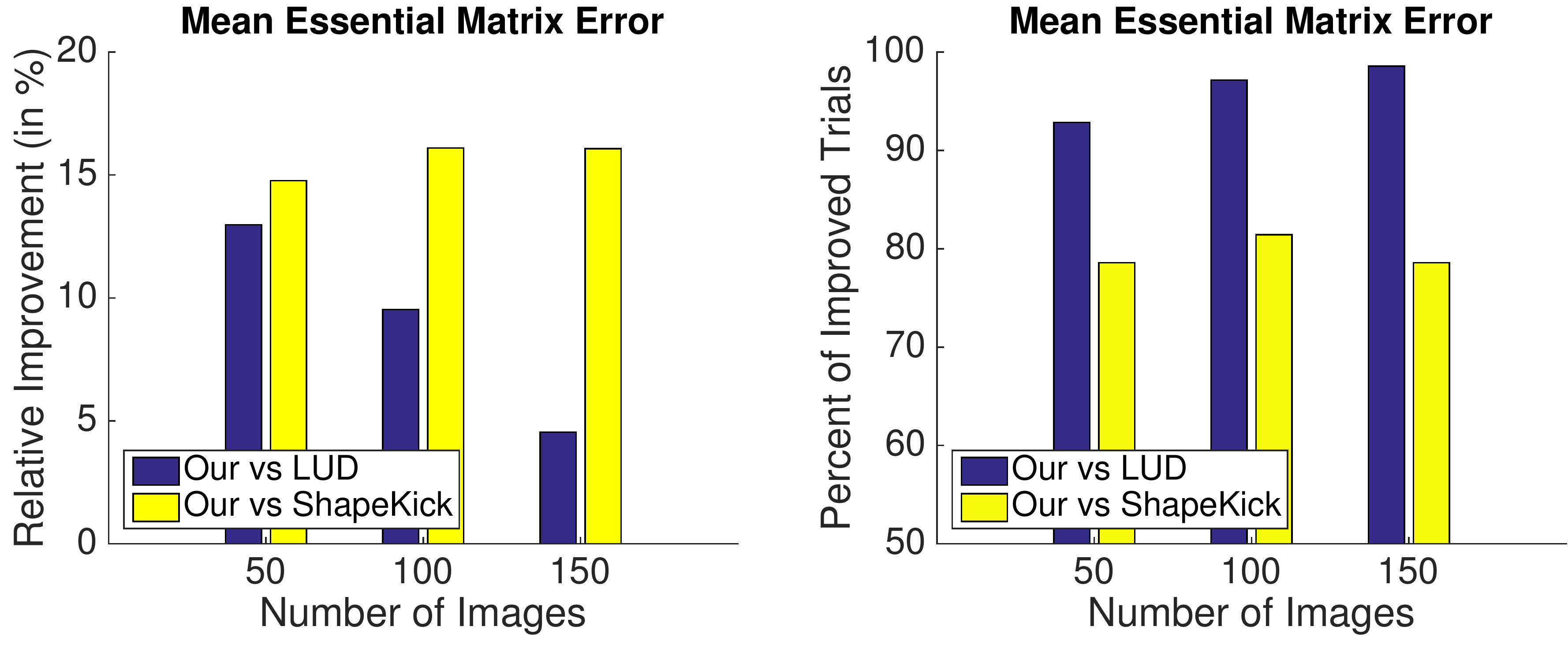}
		\label{fig:E}
	\end{subfigure}
	\begin{subfigure}{0.48\textwidth}
		\includegraphics[width=\textwidth]{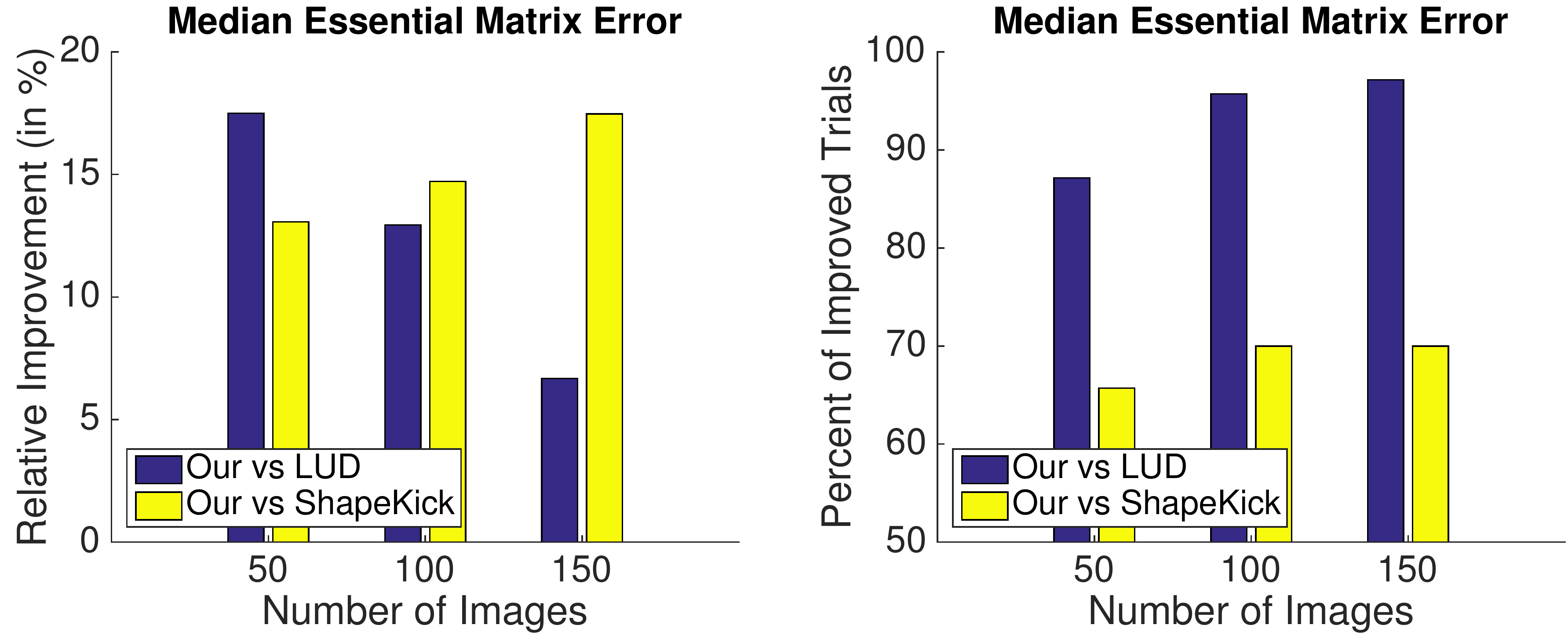}
		\label{fig:E_mi}
	\end{subfigure}
	\caption{\small These graphs show a comparison of the recovery error of essential matrices achieved with our method compared to LUD (in blue) and ShapeKick (in yellow), for collections of 50, 100, and 150 images from \cite{wilson2014robust}, The graphs on the left show the amount of relative improvement and the ones on the right show the fraction of improved trials.}
	\label{fig:result-ess}
\end{figure}
\begin{figure}[!h]
	\centering
	\begin{subfigure}{0.48\textwidth}
		\includegraphics[width=\textwidth]{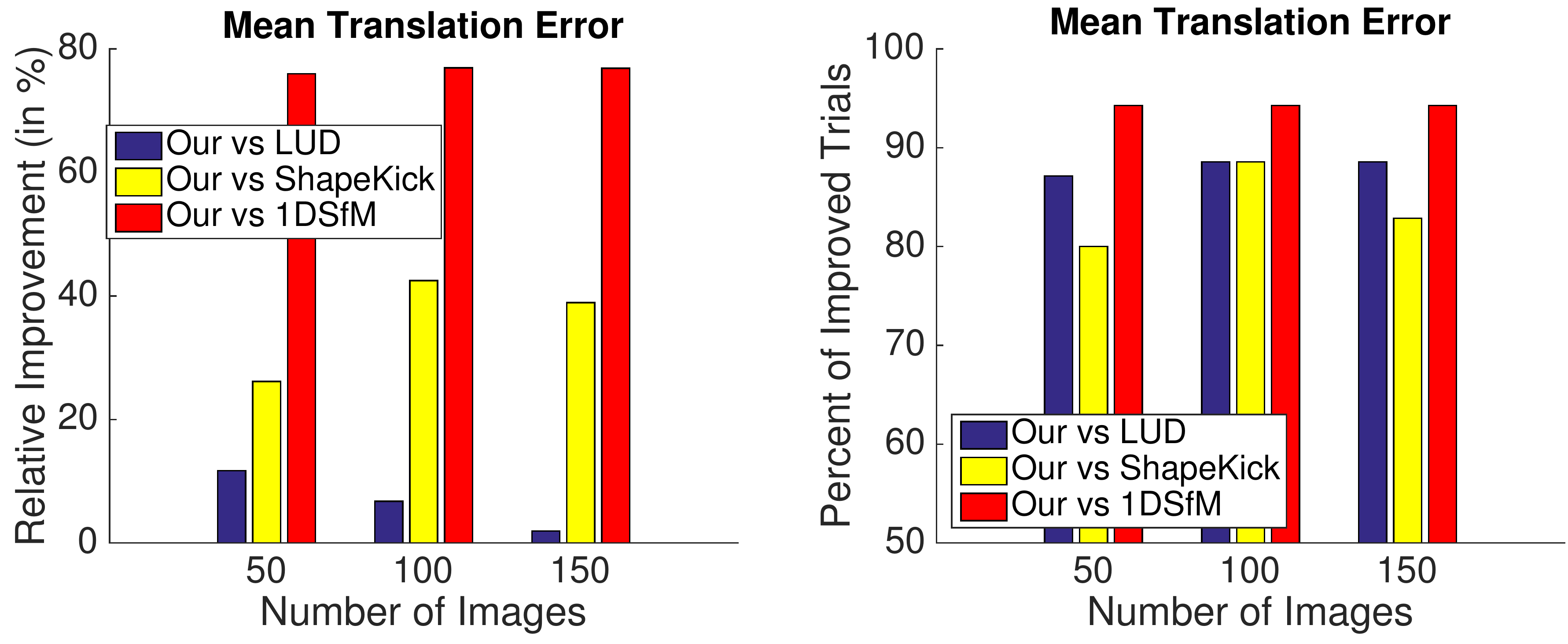}
		\label{fig:T}
	\end{subfigure}
	\begin{subfigure}{0.48\textwidth}
		\includegraphics[width=\textwidth]{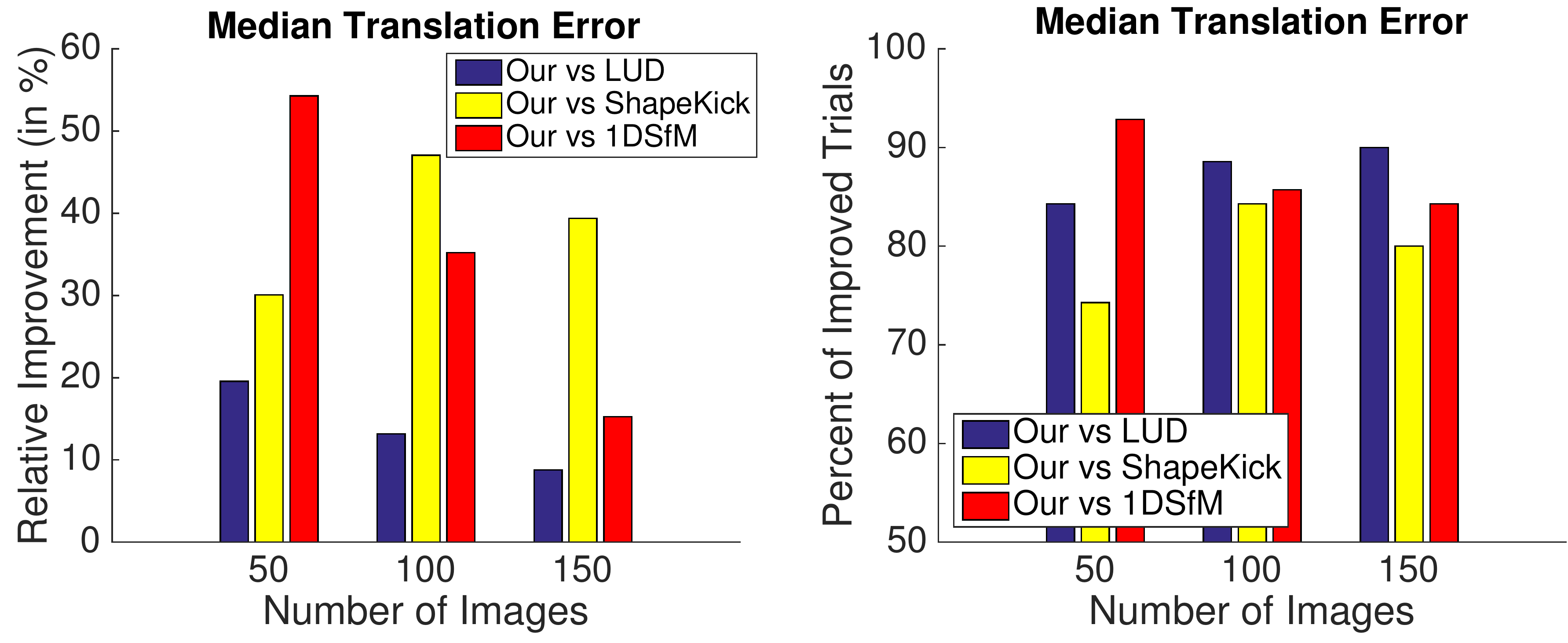}
		\label{fig:T_mi}
	\end{subfigure}
	\caption{\small A comparison of the recovery error of camera locations achieved with our method compared to LUD (in blue) and ShapeKick (in yellow), and 1DSfM (in red) for collections of 50, 100, and 150 images from \cite{wilson2014robust}, The graphs on the left show the amount of relative improvement and the ones on the right show the fraction of improved trials.}
	\label{fig:result-loc}
\end{figure}

\begin{figure*}[!h]
	\centering
	\includegraphics[width=1\textwidth]{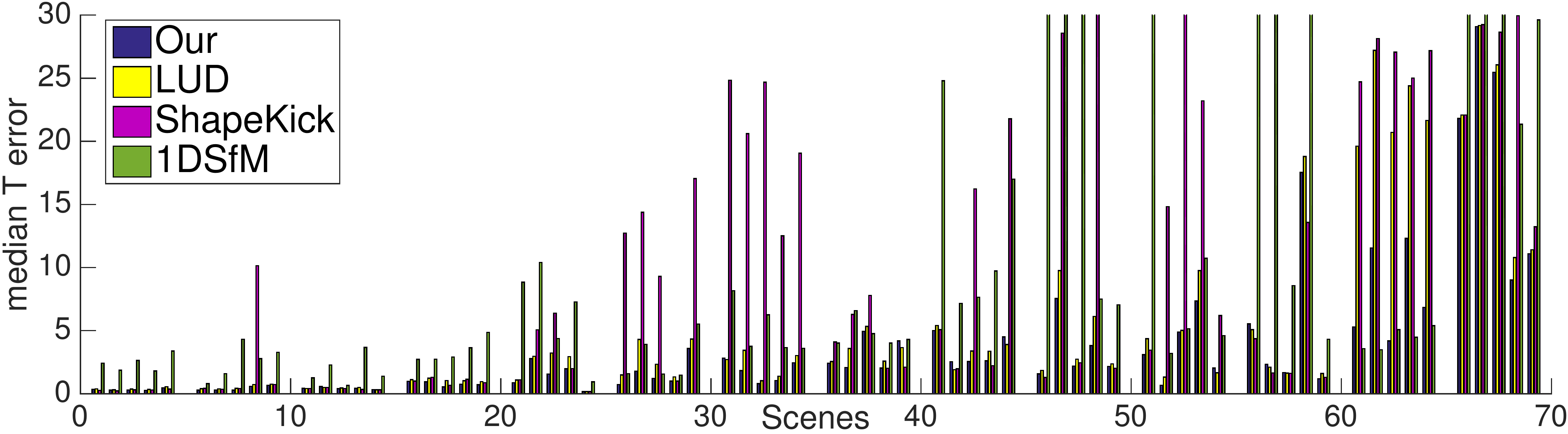}
	\caption{\small Median camera location error obtained by the four algorithms for 5 subsets of 50 images for 14 different scenes (`Notre Dame', `Montreal Notre Dame', `Alamo', `Piazza del Popolo', `Piccadilly', `NYC Library', `Yorkminster', `Union Square', `Madrid Metropolis', `Tower of London', `Vienna Cathedral', `Roman Forum' and `Ellis Island', `Gendarmenmarkt'). For clarity we terminate the median T error axis at 30.}
	\label{fig:err}	
\end{figure*}

Figures \ref{fig:result-ess}-\ref{fig:result-loc} show our results. Each graph summarizes the results of 70 trials with each value of $N$. Figure \ref{fig:result-ess} shows the quality of our essential matrix estimates compared to those obtained with LUD and ShapeKick, and Figure \ref{fig:result-loc} shows the quality of our camera location estimates compared to those achieved by the three competing algorithms. We measure these as follows. In each experiment $k$ we consider the collection of pairwise essential matrices produced by our method. We first normalize each matrix and measure its error to the respective (normalized) ground truth matrix. We then take the mean (or median) of this error over all essential matrices. Denote this error by $e^{\mathrm{Our}}_k$. We then produce similar error measures for each competing algorithm, denoted $e^{\mathrm{Other}}_k$. We then report:
\begin{itemize}
\item \textbf{Relative Improvement (in $\%$)}: Here we report for each N and competing algorithm the average of $\frac{e^{\mathrm{Other}}_k - e^{\mathrm{Our}}_k}{e^{\mathrm{Other}}_k}$ over all experiments.
\item \textbf{Percent of Improved Trials}: This provides the percentage of trials in which our algorithm achieved more accurate results than a competing algorithm, i.e., $\frac{1}{K}\sum_{k=1}^K \mathbb{I}(e^{\mathrm{Our}}_k < e^{\mathrm{Other}}_k)$, where $\mathbb{I}(.)$ is the indicator function and $K$ denotes the total number of trials.
\end{itemize}

We provide similar measures to assess the quality of our camera locations estimates. In Figure~\ref{fig:err} we further show the median error of camera location estimates for all methods in all trials for $N=50$.

It can be seen overall that our method leads to improved estimation of essential matrices and of camera locations. With 50 images, compared to, e.g., LUD, our algorithm improves the median essential matrix estimates by 17.69\%. With 150 images a smaller overall improvement of 6.68\% is achieved. This suggests that our constraints are more effective when smaller numbers of images are used. Interestingly, however, despite this reduction the fraction of trials in which our method achieved more accurate estimates compared to LUD in fact increased slightly from 87\% with 50 images to 98\% with 150 images, indicating that our method remains effective also with larger number of images (albeit yielding smaller improvement). Similar results are observed for camera location estimation. With 50 and 150 images our algorithms improves the median camera location error by 19.73\% and 8.77\% respectively, while the fraction of trials in which our method achieved more accurate estimates than LUD increased slightly from 84\% with 50 images to 90\% with 150 images.

\begin{figure}[!h]
	\centering
	\includegraphics[width=0.35\textwidth]{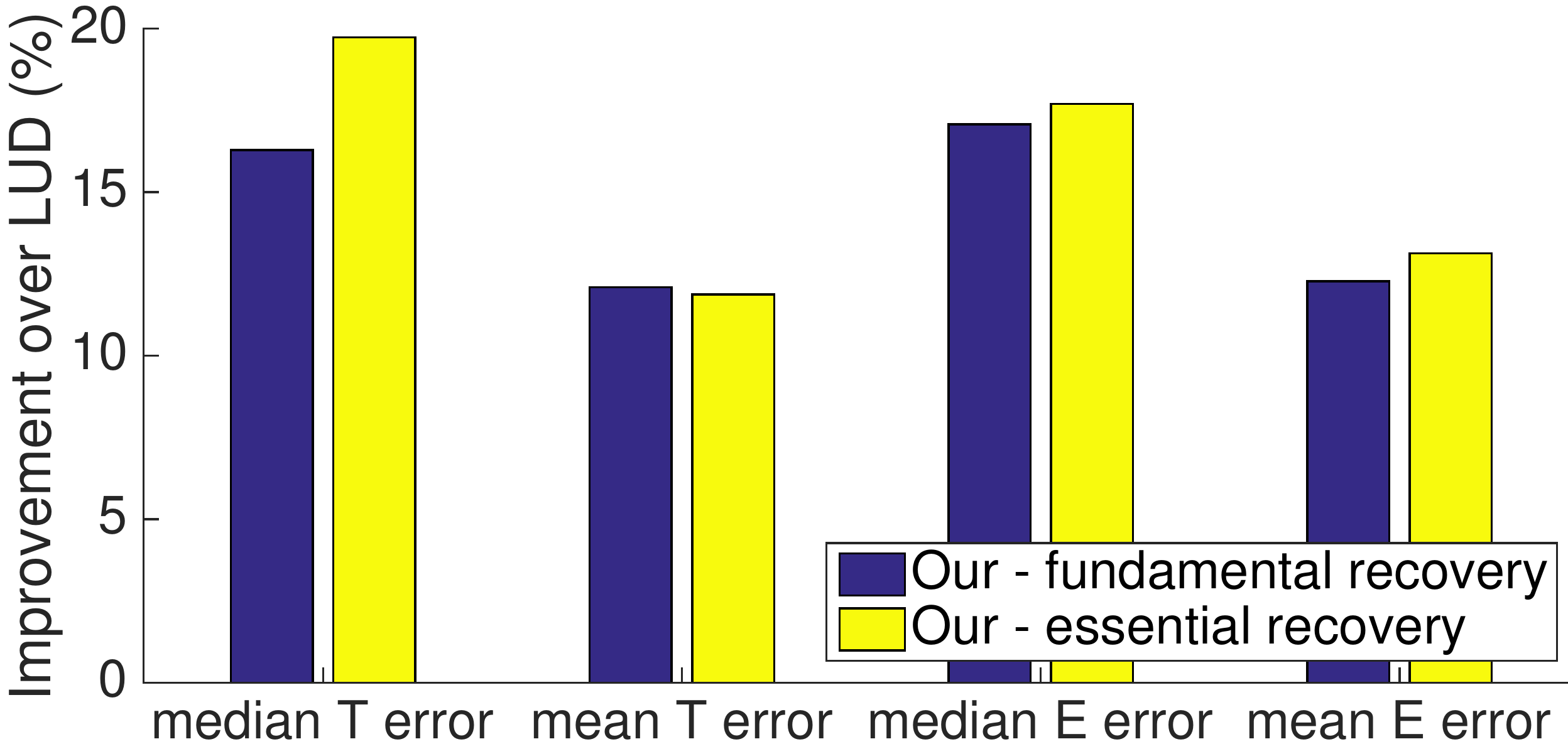}
	\caption{\small Improvement of our method over LUD using fundamental matrix (in blue) and essential matrix (yellow) for 50 images.}
	\label{fig:fund}	
\end{figure}

In our previous experiments we applied our optimization algorithm to essential matrices, assuming calibration is given. Below we further apply our algorithm to fundamental matrices in an uncalibrated setting. Since not all the entries of a $3 \times 3$ fundamental matrix are of same orders of magnitude, we normalize each of the input pairwise fundamental matrices by centering all the images and scaling them uniformly to within the $[−1, 1]$ square and then compute a normalized fundamental matrix. This does not affect our rank constraint and can be inverted at the end of the process. We tested our method on 5 subsamples of 50 images for 14 different scenes and compared it to LUD. To evaluate the quality of the recovered fundamental matrices we convert them to essential matrices by applying the known calibration matrices and further use these to recover camera locations. The results can be seen in Figure 6. Using our method to recover fundamentals (in blue) yielded comparable accuracies to our results for essential matrix recovery (yellow) and both our approaches improve significantly (10-20\%) over LUD as shown in Figure \ref{fig:fund}.


We further performed bundle adjustment (using \cite{lourakis2009sba}) initialized by the camera parameters obtained with our method and LUD. After bundle adjustment compared to LUD our method improved camera location estimates on average by 11.52\%, 3.13\% and 5.43\%, improving in 70.59\%, 64.29\% and 63.77\% of all trials for 50, 100 and 150 images respectively in terms of median translation error. These results indicate that our method maintains improved accuracies over LUD also after bundle adjustment.

With 50 images the recovery of essential matrices with our method requires roughly 20 iterations of IRLS and 1000 iterations of ADMM. These take overall about 2 minute on a 2.7 GHz Intel Core i5 computer.

To conclude, these experiments indicate that our characterization of essential matrices in multiview settings can be used to improve essential matrix and cameral location estimates. The advantage of these constraints appear to be particularly pronounced when fewer images are available.

\section{Conclusion}

We have introduced in this paper novel rank constraints on fundamental matrices in multiview settings. We have shown in particular that with non-collinear cameras the matrix that depicts the pairwise fundamentals is of rank 6 and forms the symmetric part of a rank 3 matrix whose factors are related directly to the entries of the respective camera matrices. We have used these constraints to develop an optimization framework to efficiently recover fundamental matrices for all pairs of images and to estimate their proper scale factors. Our experiments indicate that our method is able to provide improved estimates of essential matrices and camera locations in global SfM settings. Moreover, these experiments suggest that our constraints are particularly useful when fewer images are available.

Our plans for the future include improving the runtime of our optimization method. We intend to explore different ways to initialize the algorithm, possibly through convex relaxations. We would further want to explore the use of our method in related applications, e.g., in camera auto-calibration.



        \bibliographystyle{ieee}
        \bibliography{sfm}



\section{Supplementary Material}

\subsection{Results on camera location error}

In Table \ref{t} we compare Our, LUD, ShapeKick and 1DSfM on 14 different scenes for $N = $ 50, 100 and 150 images. For each scene and each choice of $N$ we report the average of the median camera location error for 5 different trials. In Table \ref{t1}-\ref{t14} , each table compares four competing algorithms  on 5 different trials for 50, 100 and 150 images.

\subsection{Results on essential matrix error}
In Table \ref{e} we compare Our, LUD, ShapeKick and 1DSfM on 14 different scenes for $N = $ 50, 100 and 150 images. For each scene and each choice of $N$ we report the average of the median essential error for 5 different trials. Essential matrix error between two cameras is computed as the norm of their difference after they are normalized to unit norm. We multiply the essential matrix error by 100 and report it for convenience. In Table \ref{e1}-\ref{e14} , each table compares four competing algorithms  on 5 different trials for 50, 100 and 150 images.

\begin{table*}
	\centering	
	\caption{Average of median camera location error for 5 trials for each scene and each choice of $N$}
	\label{t}
	\resizebox{2\columnwidth}{!}{
\begin{tabular}{|c||c|c|c|c||c|c|c|c||c|c|c|c|}
		\hline
		Scene               & \multicolumn{4}{c|}{50 images}                        & \multicolumn{4}{c|}{100 images}                    & \multicolumn{4}{c|}{150 images}                                \\ \hline
		& Our            & LUD   & ShapeKick     & 1DSfM        & Our            & LUD   & ShapeKick & 1DSfM         & Our            & LUD           & ShapeKick     & 1DSfM         \\ \hline
		Alamo               & 0.43           & 0.44  & \textbf{0.39} & 1.85         & \textbf{0.47}  & 0.48  & 0.49      & 1.49          & 0.45           & 0.46          & \textbf{0.44} & 1.16          \\ \hline
		Ellis Island        & 8.03           & 22.72 & 26.42         & \textbf{4.4} & 13.8           & 21.87 & 26.61     & \textbf{3.96} & 16.45          & 20.68         & 26.41         & \textbf{3.37} \\ \hline
		Madrid Metropolis   & \textbf{3.44}  & 3.59  & 9.45          & 13.26        & \textbf{3.38}  & 3.60  & 20.62     & 10.43         & \textbf{3.01}  & 3.11          & 14.50         & 9.06          \\ \hline
		Montreal Notre Dame & \textbf{0.42}  & 0.53  & 2.41          & 2.55         & \textbf{0.50}  & 0.54  & 2.78      & 1.80          & \textbf{0.51}  & 0.54          & 2.53          & 1.50          \\ \hline
		Notre Dame          & 0.33           & 0.39  & \textbf{0.29} & 2.42         & \textbf{0.27}  & 0.30  & 0.39      & 1.91          & 0.26           & 0.29          & \textbf{0.23} & 1.42          \\ \hline
		NYC Library         & \textbf{1.66}  & 2.76  & 10.89         & 2.81         & \textbf{2.11}  & 2.58  & 14.37     & 2.12          & \textbf{1.58}  & 2.02          & 13.85         & 2.33          \\
\hline
	\end{tabular}}
\end{table*}

\begin{table*}[!h]
	\centering
	\caption{Average of median essential matrix error for 5 trials for each scene and each choice of $N$}
	\label{e}
	\resizebox{2\columnwidth}{!}{\begin{tabular}{|c|c|c|c|c|c|c|c|c|c|}
			\hline
			Scene               & \multicolumn{3}{c|}{50 images}          & \multicolumn{3}{c|}{100 images}         & \multicolumn{3}{c|}{150 images}                 \\ \hline
			& Our            & LUD   & ShapeKick      & Our            & LUD   & ShapeKick      & Our            & LUD            & ShapeKick     \\ \hline
			Alamo               & 3.96           & 3.90  & \textbf{3.74}  & \textbf{4.77}  & 4.86  & 5.19           & \textbf{4.52}  & 4.57           & 4.64          \\ \hline
			Ellis Island        & \textbf{30.34} & 72.05 & 82.64          & \textbf{41.66} & 69.42 & 81.31          & \textbf{48.87} & 66.58          & 78.28         \\ \hline
			Madrid Metropolis   & 11.69          & 12.18 & \textbf{10.48} & 12.68          & 13.09 & \textbf{10.72} & \textbf{13.11} & 13.48          & 16.30         \\ \hline
			Montreal Notre Dame & 17.95          & 19.67 & \textbf{17.34} & \textbf{5.21}  & 5.53  & 5.86           & \textbf{4.92}  & 5.14           & 5.69          \\ \hline
			Notre Dame          & 4.90           & 5.07  & \textbf{4.37}  & \textbf{3.86}  & 4.18  & 4.44           & 3.71           & 3.89           & \textbf{3.52} \\ \hline
			NYC Library         & \textbf{19.63} & 24.84 & 21.73          & \textbf{20.37} & 24.32 & 29.68          & \textbf{19.21} & 21.28          & 34.84         \\
\hline
		\end{tabular}}
	\end{table*}

\begin{table*}[]
	\centering
	\caption{Median translation error on `Alamo'}
	\label{t1}
	\resizebox{2\columnwidth}{!}{
	\begin{tabular}{|c|c|c|c|c|c|c|c|c|c|c|c|c|c|c|c|}
		\hline
		trials    & \multicolumn{5}{c|}{50 images}                                                & \multicolumn{5}{c|}{100 images}                                               & \multicolumn{5}{c|}{150 images}                                               \\ \hline
		& 1             & 2             & 3             & 4             & 5             & 1             & 2             & 3             & 4             & 5             & 1             & 2             & 3             & 4             & 5             \\ \hline
		Our       & 0.43          & 0.56          & \textbf{0.39} & 0.42          & 0.33          & 0.45          & \textbf{0.40} & \textbf{0.39} & \textbf{0.36} & \textbf{0.75} & 0.43          & \textbf{0.37} & \textbf{0.39} & \textbf{0.39} & 0.64          \\ \hline
		LUD       & \textbf{0.41} & 0.50          & 0.45          & 0.50          & \textbf{0.32} & 0.46          & 0.41          & \textbf{0.39} & 0.38          & 0.76          & 0.43          & 0.38          & 0.40          & 0.41          & 0.69          \\ \hline
		ShapeKick & \textbf{0.41} & \textbf{0.49} & 0.41          & \textbf{0.34} & \textbf{0.32} & \textbf{0.43} & 0.41          & 0.43          & \textbf{0.36} & 0.82          & \textbf{0.42} & 0.38          & 0.40          & \textbf{0.39} & \textbf{0.63} \\ \hline
		1DSfM     & 1.26          & 2.28          & 0.65          & 3.67          & 1.39          & 1.09          & 1.12          & 0.82          & 0.74          & 3.68          & 1.04          & 0.80          & 0.62          & 0.82          & 2.53          \\ \hline
	\end{tabular}}
\end{table*}

\begin{table*}[]
	\centering
	\caption{Median translation error on `Ellis Island'}
	\label{t2}
	\resizebox{2\columnwidth}{!}{\begin{tabular}{|c|c|c|c|c|c|c|c|c|c|c|c|c|c|c|c|}
		\hline
		trials    & \multicolumn{5}{c|}{50 images}                                                & \multicolumn{5}{c|}{100 images}                                               & \multicolumn{5}{c|}{150 images}                                               \\ \hline
		& 1             & 2             & 3             & 4             & 5             & 1             & 2             & 3             & 4             & 5             & 1             & 2             & 3             & 4             & 5             \\ \hline
		Our       & 5.28          & 11.54         & \textbf{4.19} & 12.31         & 6.84          & 12.19         & 17.20         & 12.87         & 16.53         & 10.20         & 18.22         & 16.20         & 15.03         & 17.24         & 15.56         \\ \hline
		LUD       & 19.62         & 27.20         & 20.70         & 24.40         & 21.66         & 18.00         & 24.81         & 20.36         & 27.02         & 19.17         & 21.89         & 21.48         & 17.55         & 23.49         & 18.98         \\ \hline
		ShapeKick & 24.72         & 28.15         & 27.06         & 25.02         & 27.18         & 26.09         & 26.85         & 26.40         & 27.22         & 26.50         & 26.93         & 26.76         & 25.75         & 26.46         & 26.13         \\ \hline
		1DSfM     & \textbf{3.57} & \textbf{3.48} & 5.07          & \textbf{4.47} & \textbf{5.39} & \textbf{3.33} & \textbf{4.17} & \textbf{3.94} & \textbf{3.89} & \textbf{4.47} & \textbf{3.26} & \textbf{3.20} & \textbf{3.94} & \textbf{2.84} & \textbf{3.60} \\ \hline
	\end{tabular}}
\end{table*}

\begin{table*}[]
	\centering
	\caption{Median translation error on `Madrid Metropolis'}
	\label{t3}
	\resizebox{2\columnwidth}{!}{\begin{tabular}{|c|c|c|c|c|c|c|c|c|c|c|c|c|c|c|c|}
		\hline
		trials    & \multicolumn{5}{c|}{50 images}                                                & \multicolumn{5}{c|}{100 images}                                               & \multicolumn{5}{c|}{150 images}                                               \\ \hline
		& 1             & 2             & 3             & 4             & 5             & 1             & 2             & 3             & 4             & 5             & 1             & 2             & 3             & 4             & 5             \\ \hline
		Our       & \textbf{5.01} & 2.53          & \textbf{2.56} & 2.60          & \textbf{4.50} & 6.19          & \textbf{1.42} & 2.12          & \textbf{1.33} & \textbf{5.81} & \textbf{2.71} & 3.07          & \textbf{4.19} & \textbf{1.54} & \textbf{3.54} \\ \hline
		LUD       & 5.40          & \textbf{1.90} & 3.39          & 3.37          & 3.89          & 6.69          & 1.67          & \textbf{2.05} & 1.51          & 6.08          & 2.96          & 3.09          & 4.21          & 1.63          & 3.66          \\ \hline
		ShapeKick & 5.07          & 1.97          & 16.22         & \textbf{2.22} & 21.78         & \textbf{5.07} & 29.63         & 16.90         & 29.38         & 22.13         & 6.56          & \textbf{2.88} & 16.91         & 28.57         & 17.57         \\ \hline
		1DSfM     & 24.79         & 7.15          & 7.64          & 9.73          & 17.00         & 20.38         & 6.38          & 6.99          & 6.31          & 12.07         & 12.69         & 9.74          & 6.71          & 6.71          & 9.43          \\ \hline
	\end{tabular}}
\end{table*}

\begin{table*}[]
	\centering
	\caption{Median translation error on `Montreal Notre Dame'}
	\label{t4}
	\resizebox{2\columnwidth}{!}{\begin{tabular}{|c|c|c|c|c|c|c|c|c|c|c|c|c|c|c|c|}
		\hline
		trials    & \multicolumn{5}{c|}{50 images}                                                & \multicolumn{5}{c|}{100 images}                                               & \multicolumn{5}{c|}{150 images}                                               \\ \hline
		& 1             & 2             & 3             & 4             & 5             & 1             & 2             & 3             & 4             & 5             & 1             & 2             & 3             & 4             & 5             \\ \hline
		Our       & \textbf{0.28} & \textbf{0.29} & \textbf{0.28} & \textbf{0.58} & \textbf{0.67} & \textbf{0.44} & \textbf{0.36} & \textbf{0.62} & \textbf{0.47} & \textbf{0.57} & \textbf{0.45} & \textbf{0.40} & \textbf{0.72} & \textbf{0.47} & \textbf{0.53} \\ \hline
		LUD       & 0.39          & 0.39          & 0.42          & 0.72          & 0.75          & 0.45          & 0.39          & 0.70          & 0.55          & 0.61          & 0.47          & 0.41          & 0.74          & 0.51          & 0.56          \\ \hline
		ShapeKick & 0.42          & 0.36          & 0.41          & 10.14         & 0.72          & 0.46          & 0.41          & 1.50          & 10.88         & 0.64          & \textbf{0.45} & 10.22         & 0.77          & 0.61          & 0.58          \\ \hline
		1DSfM     & 0.79          & 1.59          & 4.32          & 2.79          & 3.28          & 1.67          & 2.42          & 2.19          & 1.12          & 1.58          & 1.45          & 1.65          & 2.43          & 1.20          & 0.75          \\ \hline
	\end{tabular}}
\end{table*}

\begin{table*}[]
	\centering
	\caption{Median translation error on `Notre Dame'}
	\label{t5}
	\resizebox{2\columnwidth}{!}{\begin{tabular}{|c|c|c|c|c|c|c|c|c|c|c|c|c|c|c|c|}
		\hline
		trials    & \multicolumn{5}{c|}{50 images}                                                & \multicolumn{5}{c|}{100 images}                                               & \multicolumn{5}{c|}{150 images}                                               \\ \hline
		& 1             & 2             & 3             & 4             & 5             & 1             & 2             & 3             & 4             & 5             & 1             & 2             & 3             & 4             & 5             \\ \hline
		Our       & 0.36          & 0.28          & \textbf{0.29} & \textbf{0.27} & 0.47          & 0.29          & \textbf{0.23} & \textbf{0.27} & \textbf{0.29} & \textbf{0.26} & 0.26          & 0.23          & 0.27          & 0.31          & 0.25          \\ \hline
		LUD       & 0.38          & 0.31          & 0.37          & 0.34          & 0.56          & 0.31          & 0.27          & \textbf{0.27} & 0.32          & 0.32          & 0.26          & 0.25          & 0.29          & 0.33          & 0.30          \\ \hline
		ShapeKick & \textbf{0.26} & \textbf{0.23} & 0.31          & 0.30          & \textbf{0.36} & \textbf{0.24} & 0.48          & 0.33          & 0.33          & 0.58          & \textbf{0.21} & \textbf{0.20} & \textbf{0.24} & \textbf{0.28} & \textbf{0.20} \\ \hline
		1DSfM     & 2.42          & 1.87          & 2.64          & 1.81          & 3.38          & 2.24          & 1.50          & 1.17          & 2.58          & 2.06          & 2.70          & 1.12          & 0.80          & 0.79          & 1.67          \\ \hline
	\end{tabular}}
\end{table*}

\begin{table*}[]
	\centering
	\caption{Median translation error on `NYC Library'}
	\label{t6}
	\resizebox{2\columnwidth}{!}{\begin{tabular}{|c|c|c|c|c|c|c|c|c|c|c|c|c|c|c|c|}
		\hline
		trials    & \multicolumn{5}{c|}{50 images}                                                & \multicolumn{5}{c|}{100 images}                                               & \multicolumn{5}{c|}{150 images}                                               \\ \hline
		& 1             & 2             & 3             & 4             & 5             & 1             & 2             & 3             & 4             & 5             & 1             & 2             & 3             & 4             & 5             \\ \hline
		Our       & \textbf{0.73} & \textbf{1.79} & \textbf{1.21} & \textbf{0.99} & \textbf{3.58} & \textbf{1.28} & 1.45          & 3.03          & \textbf{1.49} & 3.29          & \textbf{1.34} & \textbf{1.24} & \textbf{2.26} & \textbf{1.19} & \textbf{1.88} \\ \hline
		LUD       & 1.49          & 4.30          & 2.34          & 1.34          & 4.33          & 1.95          & 2.23          & 3.23          & 1.99          & 3.50          & 1.86          & 1.80          & 2.57          & 1.61          & 2.26          \\ \hline
		ShapeKick & 12.73         & 14.38         & 9.29          & 1.00          & 17.04         & 14.20         & 15.10         & 13.76         & 14.30         & 14.50         & 13.19         & 14.26         & 13.61         & 14.24         & 13.93         \\ \hline
		1DSfM     & 1.59          & 3.91          & 1.56          & 1.48          & 5.51          & 1.64          & \textbf{1.41} & \textbf{2.62} & 1.78          & \textbf{3.12} & 2.73          & 1.58          & 2.67          & 1.55          & 3.13          \\ \hline
	\end{tabular}}
\end{table*}

\begin{table*}[]
	\centering
	\caption{Piazza Del Popolo}
	\label{t7}
	\resizebox{2\columnwidth}{!}{\begin{tabular}{|c|c|c|c|c|c|c|c|c|c|c|c|c|c|c|c|}
		\hline
		trials    & \multicolumn{5}{c|}{50 images}                                                & \multicolumn{5}{c|}{100 images}                                               & \multicolumn{5}{c|}{150 images}                                               \\ \hline
		& 1             & 2             & 3             & 4             & 5             & 1             & 2             & 3             & 4             & 5             & 1             & 2             & 3             & 4             & 5             \\ \hline
		Our       & \textbf{0.99} & \textbf{0.95} & \textbf{0.55} & \textbf{0.75} & \textbf{0.72} & 1.39          & \textbf{1.11} & \textbf{0.72} & \textbf{1.11} & \textbf{1.74} & \textbf{1.67} & \textbf{1.57} & 1.58          & 1.49          & 1.89          \\ \hline
		LUD       & 1.12          & 1.23          & 1.04          & 1.03          & 0.96          & 1.41          & 1.28          & 0.92          & 1.68          & 2.00          & 1.80          & 1.62          & 1.64          & 1.59          & 1.97          \\ \hline
		ShapeKick & 1.00          & 1.30          & 0.67          & 1.14          & 0.87          & \textbf{1.27} & 13.35         & 0.85          & 14.64         & 18.54         & 16.79         & 18.36         & \textbf{1.23} & \textbf{1.44} & \textbf{1.72} \\ \hline
		1DSfM     & 2.72          & 2.74          & 2.90          & 3.64          & 4.85          & 2.77          & 1.77          & 1.26          & 2.85          & 3.77          & 3.97          & 2.62          & 1.85          & 3.33          & 2.05          \\ \hline
	\end{tabular}}
\end{table*}

\begin{table*}[]
	\centering
	\caption{Median translation error on `Piccadilly'}
	\label{t8}
	\resizebox{2\columnwidth}{!}{\begin{tabular}{|c|c|c|c|c|c|c|c|c|c|c|c|c|c|c|c|}
		\hline
		trials    & \multicolumn{5}{c|}{50 images}                                                & \multicolumn{5}{c|}{100 images}                                               & \multicolumn{5}{c|}{150 images}                                               \\ \hline
		& 1             & 2             & 3             & 4             & 5             & 1             & 2             & 3             & 4             & 5             & 1             & 2             & 3             & 4             & 5             \\ \hline
		Our       & \textbf{0.87} & \textbf{2.78} & \textbf{1.56} & \textbf{1.98} & \textbf{0.16} & \textbf{2.17} & \textbf{1.71} & \textbf{2.81} & \textbf{2.35} & \textbf{0.22} & \textbf{1.73} & \textbf{1.82} & 1.97          & \textbf{2.45} & 3.08          \\ \hline
		LUD       & 1.08          & 2.97          & 3.22          & 2.93          & \textbf{0.16} & 2.18          & 2.26          & 3.27          & 3.03          & 0.28          & 1.90          & 1.88          & 2.21          & 2.83          & 3.18          \\ \hline
		ShapeKick & 1.10          & 5.07          & 6.38          & 1.99          & \textbf{0.16} & 9.28          & 10.28         & 13.09         & 9.37          & 0.31          & 9.52          & 11.18         & \textbf{1.73} & 9.80          & \textbf{0.59} \\ \hline
		1DSfM     & 8.83          & 10.40         & 4.37          & 7.26          & 0.94          & 5.56          & 7.57          & 3.69          & 5.49          & 0.53          & 4.92          & 5.39          & 3.58          & 6.10          & 3.79          \\ \hline
	\end{tabular}}
\end{table*}

\begin{table*}[]
	\centering
	\caption{Median translation error on `Roman Forum'}
	\label{t9}
	\resizebox{2\columnwidth}{!}{\begin{tabular}{|c|c|c|c|c|c|c|c|c|c|c|c|c|c|c|c|}
		\hline
		trials    & \multicolumn{5}{c|}{50 images}                                                 & \multicolumn{5}{c|}{100 images}                                               & \multicolumn{5}{c|}{150 images}                                               \\ \hline
		& 1             & 2             & 3             & 4              & 5             & 1             & 2             & 3             & 4             & 5             & 1             & 2             & 3             & 4             & 5             \\ \hline
		Our       & 5.53          & 2.33          & 1.68          & 17.53          & \textbf{1.19} & 1.07          & 3.38          & \textbf{4.33} & 9.79          & \textbf{1.71} & \textbf{3.72} & \textbf{3.84} & \textbf{2.44} & 10.02         & \textbf{2.48} \\ \hline
		LUD       & 5.08          & 2.11          & 1.63          & 18.80          & 1.60          & \textbf{0.87} & \textbf{3.29} & 4.61          & \textbf{9.76} & 1.99          & \textbf{3.72} & 3.98          & 2.94          & \textbf{8.43} & 2.52          \\ \hline
		ShapeKick & \textbf{4.36} & \textbf{1.65} & \textbf{1.62} & \textbf{13.57} & 1.30          & 0.98          & 3.49          & 37.63         & 44.24         & 20.43         & 31.57         & 40.08         & 26.75         & 44.48         & 37.56         \\ \hline
		1DSfM     & 69.33         & 56.74         & 8.56          & 62.09          & 4.31          & 28.53         & 29.28         & 23.06         & 41.96         & 10.73         & 16.21         & 6.86          & 37.05         & 38.59         & 4.34          \\ \hline
	\end{tabular}}
\end{table*}

\begin{table*}[]
	\centering
	\caption{Median translation error on `Tower of London}
	\label{t10}
	\resizebox{2\columnwidth}{!}{\begin{tabular}{|c|c|c|c|c|c|c|c|c|c|c|c|c|c|c|c|}
		\hline
		trials    & \multicolumn{5}{c|}{50 images}                                                & \multicolumn{5}{c|}{100 images}                                                & \multicolumn{5}{c|}{150 images}                                               \\ \hline
		& 1             & 2             & 3             & 4             & 5             & 1              & 2             & 3             & 4             & 5             & 1             & 2             & 3             & 4             & 5             \\ \hline
		Our       & 1.57          & \textbf{7.55} & \textbf{2.19} & \textbf{3.81} & 2.16          & \textbf{15.43} & 4.29          & 7.18          & \textbf{6.01} & \textbf{2.01} & \textbf{6.15} & 4.40          & \textbf{3.88} & \textbf{4.10} & \textbf{2.52} \\ \hline
		LUD       & 1.85          & 9.75          & 2.74          & 6.13          & 2.36          & 15.91          & 4.37          & 7.70          & 6.44          & 2.34          & 6.28          & 4.61          & 4.01          & 4.41          & 2.80          \\ \hline
		ShapeKick & \textbf{1.29} & 28.58         & 2.44          & 52.95         & \textbf{2.02} & 62.18          & \textbf{2.92} & \textbf{3.55} & 61.53         & 68.28         & 59.57         & \textbf{2.53} & 67.39         & 55.23         & 63.94         \\ \hline
		1DSfM     & 127.25        & 38.86         & 61.06         & 7.51          & 7.05          & 44.91          & 21.32         & 25.77         & 15.51         & 4.70          & 48.37         & 13.50         & 9.83          & 7.82          & 7.92          \\ \hline
	\end{tabular}}
\end{table*}

\begin{table*}[]
	\centering
	\caption{Median translation error on `Union Square'}
	\label{t11}
	\resizebox{2\columnwidth}{!}{\begin{tabular}{|c|c|c|c|c|c|c|c|c|c|c|c|c|c|c|c|}
		\hline
		trials    & \multicolumn{5}{c|}{50 images}                                                & \multicolumn{5}{c|}{100 images}                                               & \multicolumn{5}{c|}{150 images}                                               \\ \hline
		& 1             & 2             & 3             & 4             & 5             & 1             & 2             & 3             & 4             & 5             & 1             & 2             & 3             & 4             & 5             \\ \hline
		Our       & \textbf{2.42} & \textbf{2.08} & 4.95          & 2.03          & 4.19          & \textbf{7.46} & \textbf{3.79} & \textbf{1.19} & 4.14          & \textbf{5.72} & 8.03          & 10.76         & 6.53          & \textbf{5.27} & 7.29          \\ \hline
		LUD       & 2.54          & 3.58          & 5.34          & 2.58          & 3.65          & 7.96          & 4.53          & 1.46          & 4.08          & 4.59          & 7.62          & 11.56         & \textbf{6.41} & 9.91          & 7.78          \\ \hline
		ShapeKick & 4.10          & 6.29          & 7.79          & \textbf{2.02} & \textbf{2.11} & 8.15          & 6.18          & 1.44          & \textbf{2.59} & 8.48          & 11.21         & 14.64         & 6.96          & 15.48         & 12.04         \\ \hline
		1DSfM     & 4.01          & 6.58          & \textbf{4.76} & 4.01          & 4.31          & 10.34         & 15.75         & 3.60          & 15.12         & 11.03         & \textbf{5.67} & \textbf{9.83} & 15.24         & 17.63         & \textbf{4.85} \\ \hline
	\end{tabular}}
\end{table*}

\begin{table*}[]
	\centering
	\caption{Median translation error on `Vienna Cathedral'}
	\label{t12}
	\resizebox{2\columnwidth}{!}{\begin{tabular}{|c|c|c|c|c|c|c|c|c|c|c|c|c|c|c|c|}
		\hline
		trials    & \multicolumn{5}{c|}{50 images}                                                & \multicolumn{5}{c|}{100 images}                                               & \multicolumn{5}{c|}{150 images}                                               \\ \hline
		& 1             & 2             & 3             & 4             & 5             & 1             & 2             & 3             & 4             & 5             & 1             & 2             & 3             & 4             & 5             \\ \hline
		Our       & \textbf{3.09} & \textbf{0.67} & \textbf{4.88} & \textbf{7.35} & 2.04          & \textbf{6.23} & \textbf{0.99} & 4.23          & \textbf{3.33} & 4.67          & 8.77          & \textbf{1.49} & \textbf{5.00} & 4.43          & \textbf{6.14} \\ \hline
		LUD       & 4.36          & 1.31          & 5.01          & 9.75          & \textbf{1.66} & 8.57          & 1.52          & 5.47          & 5.91          & 6.37          & 8.94          & 2.01          & 5.93          & 5.47          & 7.10          \\ \hline
		ShapeKick & 3.45          & 14.81         & 31.26         & 23.21         & 6.20          & 25.26         & 19.73         & 23.83         & 32.15         & 28.11         & 24.27         & 19.34         & 30.71         & 22.89         & 25.72         \\ \hline
		1DSfM     & 67.82         & 3.20          & 5.15          & 10.74         & 4.60          & 6.27          & 4.54          & \textbf{4.11} & 6.41          & \textbf{4.49} & \textbf{4.83} & 2.37          & 5.41          & \textbf{4.12} & 6.89          \\ \hline
	\end{tabular}}
\end{table*}

\begin{table*}[]
	\centering
	\caption{Median translation error on `Yorkminster'}
	\label{t13}
	\resizebox{2\columnwidth}{!}{\begin{tabular}{|c|c|c|c|c|c|c|c|c|c|c|c|c|c|c|c|}
		\hline
		trials    & \multicolumn{5}{c|}{50 images}                                                & \multicolumn{5}{c|}{100 images}                                               & \multicolumn{5}{c|}{150 images}                                               \\ \hline
		& 1             & 2             & 3             & 4             & 5             & 1             & 2             & 3             & 4             & 5             & 1             & 2             & 3             & 4             & 5             \\ \hline
		Our       & 2.82          & \textbf{1.83} & \textbf{0.79} & \textbf{1.04} & \textbf{2.45} & \textbf{1.46} & \textbf{1.88} & \textbf{3.17} & \textbf{1.65} & \textbf{2.83} & \textbf{2.54} & \textbf{2.22} & \textbf{3.17} & \textbf{1.66} & \textbf{2.98} \\ \hline
		LUD       & \textbf{2.71} & 3.44          & 1.02          & 1.38          & 3.02          & 1.82          & 2.36          & 3.41          & 2.04          & 3.14          & 2.64          & 2.33          & 3.30          & 1.83          & 3.30          \\ \hline
		ShapeKick & 24.83         & 20.61         & 24.68         & 12.51         & 19.05         & 17.15         & 15.19         & 23.57         & 15.94         & 21.51         & 17.32         & 16.95         & 23.01         & 14.93         & 21.61         \\ \hline
		1DSfM     & 8.17          & 3.76          & 6.25          & 3.66          & 3.59          & 8.32          & 6.39          & 7.07          & 5.94          & 6.58          & 8.87          & 6.52          & 6.89          & 5.86          & 4.05          \\ \hline
	\end{tabular}}
\end{table*}

\begin{table*}[]
	\centering
	\caption{Median translation error on `Gendarmenmarkt'}
	\label{t14}
	\resizebox{2\columnwidth}{!}{\begin{tabular}{|c|c|c|c|c|c|c|c|c|c|c|c|c|c|c|c|}
		\hline
		trials    & \multicolumn{5}{c|}{50 images}                                                    & \multicolumn{5}{c|}{100 images}                                                  & \multicolumn{5}{c|}{150 images}                                                  \\ \hline
		& 1              & 2              & 3              & 4             & 5              & 1              & 2              & 3              & 4             & 5             & 1             & 2             & 3              & 4              & 5              \\ \hline
		Our       & \textbf{21.81} & \textbf{29.08} & \textbf{25.47} & \textbf{9.00} & \textbf{11.09} & 30.80          & \textbf{19.18} & 19.63          & \textbf{7.40} & \textbf{0.69} & \textbf{4.16} & \textbf{3.23} & 19.86          & 31.45          & 27.31          \\ \hline
		LUD       & 22.08          & 29.18          & 26.05          & 10.79         & 11.40          & 31.23          & 20.51          & 19.10          & 9.90          & 0.97          & 6.09          & 4.30          & \textbf{19.69} & \textbf{31.38} & \textbf{25.82} \\ \hline
		ShapeKick & 22.08          & 29.24          & 28.65          & 29.94         & 13.23          & \textbf{30.13} & 33.03          & \textbf{12.88} & 26.88         & 1.00          & 4.62          & 3.82          & 28.96          & 32.16          & 38.57          \\ \hline
		1DSfM     & 60.17          & 38.88          & 35.73          & 21.35         & 29.61          & 45.94          & 36.53          & 21.24          & 17.12         & 55.09         & 43.35         & 36.71         & 27.86          & 49.52          & 26.09          \\ \hline
	\end{tabular}}
\end{table*}

\begin{table*}[!h]
	\centering
	\caption{Median essential matrix error on `Alamo'}
	\label{e1}
	\resizebox{2\columnwidth}{!}{\begin{tabular}{|c|c|c|c|c|c|c|c|c|c|c|c|c|c|c|c|}
		\hline
		trials    & \multicolumn{5}{c|}{50 images}                                                & \multicolumn{5}{c|}{100 images}                                               & \multicolumn{5}{c|}{150 images}                                               \\ \hline
		& 1             & 2             & 3             & 4             & 5             & 1             & 2             & 3             & 4             & 5             & 1             & 2             & 3             & 4             & 5             \\ \hline
		Our       & 3.45          & 4.37          & 3.88          & 4.98          & 3.09          & 4.37          & \textbf{4.13} & \textbf{4.29} & \textbf{5.36} & \textbf{5.69} & 4.45          & \textbf{3.75} & \textbf{3.95} & 4.31          & \textbf{6.14} \\ \hline
		LUD       & \textbf{3.43} & 4.20          & 3.85          & 5.06          & \textbf{2.98} & 4.46          & 4.23          & 4.46          & 5.38          & 5.74          & 4.50          & 3.80          & 4.01          & 4.35          & 6.20          \\ \hline
		ShapeKick & 3.46          & \textbf{4.10} & \textbf{3.74} & \textbf{4.42} & 3.00          & \textbf{4.34} & 4.27          & 4.56          & 4.84          & 7.92          & \textbf{4.43} & 3.80          & 4.00          & \textbf{4.16} & 6.80          \\ \hline
	\end{tabular}}
\end{table*}

\clearpage

\begin{table*}[!h]
	\centering
	\caption{Median essential matrix error on `Ellis Island'}
	\label{e2}
	\resizebox{2\columnwidth}{!}{\begin{tabular}{|c|c|c|c|c|c|c|c|c|c|c|c|c|c|c|c|}
		\hline
		trials    & \multicolumn{5}{c|}{50 images}                                                     & \multicolumn{5}{c|}{100 images}                                                    & \multicolumn{5}{c|}{150 images}                                                    \\ \hline
		& 1              & 2              & 3              & 4              & 5              & 1              & 2              & 3              & 4              & 5              & 1              & 2              & 3              & 4              & 5              \\ \hline
		Our       & \textbf{21.31} & \textbf{37.73} & \textbf{21.78} & \textbf{38.26} & \textbf{32.61} & \textbf{37.81} & \textbf{43.05} & \textbf{44.16} & \textbf{47.02} & \textbf{36.27} & \textbf{51.19} & \textbf{48.04} & \textbf{47.57} & \textbf{51.98} & \textbf{45.57} \\ \hline
		LUD       & 61.47          & 78.99          & 73.37          & 80.39          & 66.01          & 67.30          & 68.53          & 71.99          & 74.29          & 64.99          & 67.83          & 66.15          & 64.62          & 70.77          & 63.50          \\ \hline
		ShapeKick & 71.34          & 86.47          & 87.20          & 91.98          & 76.20          & 79.84          & 79.59          & 84.43          & 87.27          & 75.39          & 78.52          & 77.83          & 77.06          & 83.39          & 74.61          \\ \hline
	\end{tabular}}
\end{table*}

\begin{table*}[!h]
	\centering
	\caption{Median essential matrix error on `Madrid Metropolis'}
	\label{e3}
	\resizebox{2\columnwidth}{!}{\begin{tabular}{|c|c|c|c|c|c|c|c|c|c|c|c|c|c|c|c|}
		\hline
		trials    & \multicolumn{5}{c|}{50 images}                                                   & \multicolumn{5}{c|}{100 images}                                                 & \multicolumn{5}{c|}{150 images}                                                  \\ \hline
		& 1              & 2             & 3              & 4             & 5              & 1              & 2             & 3             & 4             & 5              & 1              & 2             & 3              & 4             & 5              \\ \hline
		Our       & 15.71          & 6.36          & 18.10          & 6.08          & 12.20          & 17.65          & 6.76          & \textbf{7.99} & \textbf{5.47} & 25.54          & \textbf{11.32} & 8.13          & \textbf{23.05} & 6.95          & \textbf{16.08} \\ \hline
		LUD       & 17.11          & 6.35          & 19.33          & 5.41          & 12.71          & 17.78          & 6.87          & 8.52          & 5.59          & 26.67          & 11.52          & 8.39          & 23.62          & 7.11          & 16.79          \\ \hline
		ShapeKick & \textbf{12.59} & \textbf{5.79} & \textbf{17.46} & \textbf{4.84} & \textbf{11.71} & \textbf{10.22} & \textbf{6.09} & 8.66          & 5.79          & \textbf{22.83} & 17.87          & \textbf{7.20} & 32.42          & \textbf{6.76} & 17.25          \\ \hline
	\end{tabular}}
\end{table*}

\begin{table*}
	\centering
	\caption{Median essential matrix error on `Montreal Notre Dame'}
	\label{e4}
	\resizebox{2\columnwidth}{!}{\begin{tabular}{|c|c|c|c|c|c|c|c|c|c|c|c|c|c|c|c|}
		\hline
		trials    & \multicolumn{5}{c|}{50 images}                                                 & \multicolumn{5}{c|}{100 images}                                               & \multicolumn{5}{c|}{150 images}                                               \\ \hline
		& 1             & 2             & 3              & 4             & 5             & 1             & 2             & 3             & 4             & 5             & 1             & 2             & 3             & 4             & 5             \\ \hline
		Our       & \textbf{3.94} & \textbf{2.63} & 69.88          & \textbf{5.51} & \textbf{7.79} & \textbf{3.99} & \textbf{3.06} & \textbf{8.86} & \textbf{4.35} & \textbf{5.80} & \textbf{3.78} & \textbf{3.91} & \textbf{7.28} & \textbf{4.26} & \textbf{5.35} \\ \hline
		LUD       & 4.33          & 3.17          & 75.90          & 6.34          & 8.63          & 4.25          & 3.31          & 9.40          & 4.65          & 6.04          & 3.96          & 4.04          & 7.79          & 4.44          & 5.49          \\ \hline
		ShapeKick & 4.19          & 3.12          & \textbf{63.02} & 6.55          & 9.79          & 5.05          & 3.53          & 8.95          & 4.91          & 6.85          & 4.28          & 4.17          & 8.66          & 5.23          & 6.12          \\ \hline
	\end{tabular}}
\end{table*}

\begin{table*}[!h]
	\centering
	\caption{Median essential matrix error on `Notre Dame'}
	\label{e5}
	\resizebox{2\columnwidth}{!}{\begin{tabular}{|c|c|c|c|c|c|c|c|c|c|c|c|c|c|c|c|}
		\hline
		trials    & \multicolumn{5}{c|}{50 images}                                                & \multicolumn{5}{c|}{100 images}                                               & \multicolumn{5}{c|}{150 images}                                               \\ \hline
		& 1             & 2             & 3             & 4             & 5             & 1             & 2             & 3             & 4             & 5             & 1             & 2             & 3             & 4             & 5             \\ \hline
		Our       & 5.09          & 5.79          & 4.13          & \textbf{3.41} & 6.06          & 4.38          & \textbf{3.92} & \textbf{3.15} & \textbf{4.16} & \textbf{3.69} & 3.82          & 3.63          & 3.67          & 4.16          & 3.28          \\ \hline
		LUD       & 5.84          & 5.58          & 4.50          & 3.64          & 5.78          & 4.89          & 4.19          & 3.36          & 4.43          & 4.06          & 4.02          & 3.79          & 3.81          & 4.33          & 3.48          \\ \hline
		ShapeKick & \textbf{4.24} & \textbf{4.57} & \textbf{4.05} & 3.88          & \textbf{5.10} & \textbf{3.69} & 4.82          & 3.82          & 4.90          & 4.96          & \textbf{3.28} & \textbf{3.56} & \textbf{3.64} & \textbf{4.04} & \textbf{3.08} \\ \hline
	\end{tabular}}
\end{table*}

\begin{table*}[!h]
	\centering
	\caption{Median essential matrix error on `NYC Library'}
	\label{e6}
	\resizebox{2\columnwidth}{!}{\begin{tabular}{|c|c|c|c|c|c|c|c|c|c|c|c|c|c|c|c|}
		\hline
		trials    & \multicolumn{5}{c|}{50 images}                                                   & \multicolumn{5}{c|}{100 images}                                                    & \multicolumn{5}{c|}{150 images}                                                    \\ \hline
		& 1             & 2              & 3              & 4             & 5              & 1              & 2              & 3              & 4              & 5              & 1              & 2              & 3              & 4              & 5              \\ \hline
		Our       & 9.42          & \textbf{32.89} & 20.85          & \textbf{9.32} & \textbf{25.68} & \textbf{15.42} & \textbf{18.87} & 24.60          & \textbf{18.52} & \textbf{24.46} & \textbf{15.99} & \textbf{19.24} & \textbf{25.74} & \textbf{15.37} & \textbf{19.69} \\ \hline
		LUD       & 15.44         & 40.88          & 23.54          & 12.73         & 31.60          & 19.28          & 23.16          & 28.34          & 22.82          & 28.02          & 18.42          & 21.53          & 27.52          & 17.56          & 21.36          \\ \hline
		ShapeKick & \textbf{8.97} & 38.38          & \textbf{19.24} & 10.03         & 32.05          & 29.24          & 30.29          & \textbf{22.11} & 30.71          & 36.04          & 48.04          & 31.21          & 33.81          & 25.90          & 35.22          \\ \hline
	\end{tabular}}
\end{table*}

\begin{table*}[!h]
	\centering
	\caption{Median essential matrix error on `Piazza Del Popolo'}
	\label{e7}
	\resizebox{2\columnwidth}{!}{\begin{tabular}{|c|c|c|c|c|c|c|c|c|c|c|c|c|c|c|c|}
		\hline
		trials    & \multicolumn{5}{c|}{50 images}                                                & \multicolumn{5}{c|}{100 images}                                                & \multicolumn{5}{c|}{150 images}                                                \\ \hline
		& 1             & 2             & 3             & 4             & 5             & 1             & 2             & 3             & 4             & 5              & 1             & 2             & 3             & 4             & 5              \\ \hline
		Our       & \textbf{6.14} & \textbf{5.02} & \textbf{2.82} & \textbf{5.11} & \textbf{4.15} & 7.55          & \textbf{4.87} & \textbf{3.74} & \textbf{6.34} & \textbf{11.90} & \textbf{9.64} & \textbf{8.31} & 10.25         & 8.45          & \textbf{10.16} \\ \hline
		LUD       & 7.43          & 5.28          & 4.59          & 6.78          & 4.85          & 8.48          & 5.47          & 4.20          & 8.00          & 14.17          & 10.20         & 8.92          & 11.06         & 8.79          & 10.72          \\ \hline
		ShapeKick & 7.26          & \textbf{5.02} & 3.47          & 7.65          & 5.00          & \textbf{7.48} & 5.30          & 3.78          & 7.81          & 13.57          & 10.38         & 8.41          & \textbf{9.32} & \textbf{8.42} & 12.16          \\ \hline
	\end{tabular}}
\end{table*}

\begin{table*}[]
	\centering
	\caption{Median essential matrix error on `Piccadilly'}
	\label{e8}
	\resizebox{2\columnwidth}{!}{\begin{tabular}{|c|c|c|c|c|c|c|c|c|c|c|c|c|c|c|c|}
		\hline
		trials    & \multicolumn{5}{c|}{50 images}                                                     & \multicolumn{5}{c|}{100 images}                                                    & \multicolumn{5}{c|}{150 images}                                                    \\ \hline
		& 1              & 2              & 3              & 4              & 5              & 1              & 2              & 3              & 4              & 5              & 1              & 2              & 3              & 4              & 5              \\ \hline
		Our       & \textbf{10.32} & \textbf{16.57} & \textbf{17.33} & \textbf{24.44} & \textbf{93.02} & \textbf{16.47} & \textbf{13.24} & \textbf{19.12} & 14.33          & \textbf{81.16} & \textbf{11.83} & \textbf{11.82} & 12.58          & \textbf{15.36} & 36.19          \\ \hline
		LUD       & 10.58          & 16.80          & 25.16          & 30.89          & 94.92          & 17.76          & 14.38          & 20.07          & 17.95          & 97.95          & 12.49          & 12.29          & 13.15          & 18.08          & 36.54          \\ \hline
		ShapeKick & 10.77          & 19.83          & 37.34          & 33.28          & 94.62          & 26.25          & 18.12          & 31.36          & \textbf{13.43} & 88.60          & 17.96          & 18.86          & \textbf{12.28} & 35.14          & \textbf{29.06} \\ \hline
	\end{tabular}}
\end{table*}

\begin{table*}[]
	\centering
	\caption{Median essential matrix error on `Roman Forum'}
	\label{e9}
	\resizebox{2\columnwidth}{!}{\begin{tabular}{|c|c|c|c|c|c|c|c|c|c|c|c|c|c|c|c|}
		\hline
		trials    & \multicolumn{5}{c|}{50 images}                                                 & \multicolumn{5}{c|}{100 images}                                                 & \multicolumn{5}{c|}{150 images}                                                \\ \hline
		& 1             & 2             & 3             & 4              & 5             & 1             & 2             & 3              & 4              & 5             & 1             & 2              & 3             & 4             & 5             \\ \hline
		Our       & 11.90         & 7.15          & 6.17          & 37.92          & \textbf{5.26} & \textbf{3.00} & \textbf{7.87} & \textbf{12.43} & \textbf{11.22} & \textbf{7.16} & \textbf{7.34} & \textbf{10.07} & \textbf{7.19} & \textbf{9.65} & \textbf{6.89} \\ \hline
		LUD       & 16.52         & 8.22          & \textbf{5.38} & 43.48          & 6.56          & 5.51          & 9.31          & 13.57          & 12.18          & 7.34          & 7.58          & 10.34          & 7.57          & 10.15         & 7.12          \\ \hline
		ShapeKick & \textbf{6.91} & \textbf{5.14} & \textbf{5.38} & \textbf{25.74} & 5.68          & 4.81          & 11.20         & 15.75          & 20.56          & 10.54         & 19.81         & 33.18          & 27.84         & 29.38         & 16.91         \\ \hline
	\end{tabular}}
\end{table*}

\begin{table*}[]
	\centering
	\caption{Median essential matrix error on `Tower of London'}
	\label{e10}
	\resizebox{2\columnwidth}{!}{\begin{tabular}{|c|c|c|c|c|c|c|c|c|c|c|c|c|c|c|c|}
		\hline
		trials    & \multicolumn{5}{c|}{50 images}                                                 & \multicolumn{5}{c|}{100 images}                                                & \multicolumn{5}{c|}{150 images}                                               \\ \hline
		& 1             & 2              & 3             & 4             & 5             & 1              & 2             & 3             & 4             & 5             & 1             & 2             & 3             & 4             & 5             \\ \hline
		Our       & \textbf{3.24} & \textbf{11.29} & \textbf{4.79} & \textbf{6.90} & \textbf{3.44} & \textbf{21.79} & 8.05          & 10.02         & 8.12          & 4.82          & 8.36          & 8.25          & 7.32          & 7.08          & \textbf{5.92} \\ \hline
		LUD       & 3.42          & 12.20          & 6.86          & 10.70         & 4.17          & 22.81          & 8.34          & 10.40         & 8.60          & 5.22          & 8.51          & 8.46          & 7.58          & 7.24          & 6.14          \\ \hline
		ShapeKick & 3.28          & 29.92          & 7.93          & 12.61         & 3.86          & 38.72          & \textbf{6.99} & \textbf{6.99} & \textbf{7.95} & \textbf{4.62} & \textbf{5.24} & \textbf{6.36} & \textbf{6.63} & \textbf{5.82} & 6.27          \\ \hline
	\end{tabular}}
\end{table*}

\begin{table*}[]
	\centering
	\caption{Median essential matrix error on `Union Square'}
	\label{e11}
	\resizebox{2\columnwidth}{!}{\begin{tabular}{|c|c|c|c|c|c|c|c|c|c|c|c|c|c|c|c|}
		\hline
		trials    & \multicolumn{5}{c|}{50 images}                                                     & \multicolumn{5}{c|}{100 images}                                                    & \multicolumn{5}{c|}{150 images}                                                    \\ \hline
		& 1              & 2              & 3              & 4              & 5              & 1              & 2              & 3              & 4              & 5              & 1              & 2              & 3              & 4              & 5              \\ \hline
		Our       & 49.93          & \textbf{27.40} & \textbf{43.18} & \textbf{59.10} & 22.71          & \textbf{71.82} & \textbf{41.19} & \textbf{64.60} & 29.68          & 89.24          & \textbf{53.41} & \textbf{49.30} & 75.15          & \textbf{28.71} & \textbf{48.59} \\ \hline
		LUD       & 51.32          & 55.97          & 45.01          & 74.52          & 23.07          & 68.85          & 53.87          & 82.60          & 30.11          & \textbf{78.49} & 53.75          & 51.07          & 56.75          & 40.30          & 51.62          \\ \hline
		ShapeKick & \textbf{44.06} & 55.31          & 48.19          & 69.74          & \textbf{15.92} & 77.33          & 60.16          & 76.59          & \textbf{29.35} & 98.79          & 67.21          & 53.69          & \textbf{56.32} & 51.96          & 55.89          \\ \hline
	\end{tabular}}
\end{table*}

\begin{table*}[]
	\centering
	\caption{Median essential matrix error on `Vienna Cathedral'}
	\label{e12}
	\resizebox{2\columnwidth}{!}{\begin{tabular}{|c|c|c|c|c|c|c|c|c|c|c|c|c|c|c|c|}
		\hline
		trials    & \multicolumn{5}{c|}{50 images}                                                   & \multicolumn{5}{c|}{100 images}                                                  & \multicolumn{5}{c|}{150 images}                                                   \\ \hline
		& 1              & 2             & 3             & 4              & 5              & 1              & 2             & 3             & 4              & 5              & 1              & 2             & 3              & 4              & 5              \\ \hline
		Our       & \textbf{12.77} & \textbf{4.07} & \textbf{6.96} & \textbf{12.10} & \textbf{10.46} & \textbf{11.72} & \textbf{5.60} & \textbf{9.57} & \textbf{11.14} & \textbf{10.04} & \textbf{15.07} & \textbf{7.30} & \textbf{12.94} & \textbf{13.67} & \textbf{13.19} \\ \hline
		LUD       & 25.75          & 6.50          & 9.36          & 20.53          & 63.74          & 15.53          & 7.87          & 14.78         & 16.00          & 12.16          & 17.74          & 9.08          & 15.52          & 17.65          & 14.68          \\ \hline
		ShapeKick & 18.56          & 11.20         & 20.03         & 31.25          & 30.35          & 31.21          & 10.05         & 34.17         & 22.96          & 17.39          & 33.10          & 17.62         & 21.93          & 25.49          & 26.37          \\ \hline
	\end{tabular}}
\end{table*}

\begin{table*}[]
	\centering
	\caption{Median essential matrix error on `Yorkminster'}
	\label{e13}
	\resizebox{2\columnwidth}{!}{\begin{tabular}{|c|c|c|c|c|c|c|c|c|c|c|c|c|c|c|c|}
		\hline
		trials    & \multicolumn{5}{c|}{50 images}                                                & \multicolumn{5}{c|}{100 images}                                               & \multicolumn{5}{c|}{150 images}                                                \\ \hline
		& 1             & 2             & 3             & 4             & 5             & 1             & 2             & 3             & 4             & 5             & 1             & 2             & 3              & 4             & 5             \\ \hline
		Our       & \textbf{6.68} & \textbf{7.33} & \textbf{3.68} & \textbf{5.98} & 8.01          & 6.34          & 8.71          & 9.46          & 8.55          & 9.94          & \textbf{6.66} & 8.86          & \textbf{11.63} & 7.87          & \textbf{9.69} \\ \hline
		LUD       & 7.45          & 9.91          & 4.33          & 7.43          & 9.97          & 6.83          & 10.12         & 10.17         & 9.85          & 11.43         & 7.02          & 9.54          & 12.19          & 8.33          & 10.21         \\ \hline
		ShapeKick & 13.44         & 13.51         & 4.71          & 7.88          & \textbf{7.69} & \textbf{5.55} & \textbf{7.61} & \textbf{6.95} & \textbf{8.09} & \textbf{6.49} & 9.57          & \textbf{7.59} & 18.80          & \textbf{7.69} & 9.88          \\ \hline
	\end{tabular}}
\end{table*}

\begin{table*}[!h]
	\centering
	\caption{Median essential matrix error on `Gendermenmarkt'}
	\label{e14}
	\resizebox{2\columnwidth}{!}{\begin{tabular}{|c|c|c|c|c|c|c|c|c|c|c|c|c|c|c|c|}
		\hline
		trials    & \multicolumn{5}{c|}{50 images}                                                       & \multicolumn{5}{c|}{100 images}                                                    & \multicolumn{5}{c|}{150 images}                                                   \\ \hline
		& 1               & 2               & 3              & 4              & 5              & 1               & 2              & 3              & 4              & 5             & 1             & 2              & 3              & 4              & 5              \\ \hline
		Our       & \textbf{125.45} & \textbf{120.76} & \textbf{60.34} & \textbf{33.90} & \textbf{29.67} & 121.74          & \textbf{55.02} & 85.01          & \textbf{37.03} & \textbf{9.37} & \textbf{9.31} & 12.19          & \textbf{67.80} & \textbf{90.62} & \textbf{97.17} \\ \hline
		LUD       & 127.45          & 122.34          & 61.55          & 39.83          & 30.98          & 119.72          & 55.65          & 86.67          & 43.07          & 25.31         & 10.52         & \textbf{11.92} & 68.12          & 91.04          & 99.63          \\ \hline
		ShapeKick & 127.58          & 122.21          & 61.70          & 47.07          & 34.53          & \textbf{119.16} & 57.64          & \textbf{79.77} & 47.80          & 22.11         & 10.51         & 12.14          & 76.90          & 91.11          & 107.33         \\ \hline
	\end{tabular}}
\end{table*}

\end{document}